\documentclass[letterpaper]{article} 
\usepackage{aaai25}  
\usepackage{times}  
\usepackage{helvet}  
\usepackage{courier}  
\usepackage[hyphens]{url}  
\usepackage{graphicx} 
\usepackage{natbib}  
\usepackage{caption} 
\usepackage{algorithm}
\usepackage{algorithmic}
\usepackage{multirow}
\usepackage{amsmath}
\usepackage{pifont}
\usepackage{amssymb}
\usepackage{newfloat}
\usepackage{listings}
\usepackage{xcolor}
\usepackage{rotating}
\usepackage{newfloat}
\usepackage{listings}
\DeclareCaptionStyle{ruled}{labelfont=normalfont,labelsep=colon,strut=off} 
\floatstyle{ruled}
\newfloat{listing}{tb}{lst}{}
\floatname{listing}{Listing}
\usepackage{hyperref} 

\title{STAMPsy: Towards SpatioTemporal-Aware Mixed-Type Dialogues \\ for Psychological Counseling}
\author{
    Jieyi Wang\textsuperscript{\rm 1},
    Yue Huang\textsuperscript{\rm 1}, 
    Zeming Liu\textsuperscript{\rm 2}${}^{\ast}$, 
    Dexuan Xu\textsuperscript{\rm 1}, 
    Chuan Wang\textsuperscript{\rm 3}, 
    Xiaoming Shi\textsuperscript{\rm 4}, \\
    Ruiyuan Guan\textsuperscript{\rm 1}, 
    Hongxing Wang\textsuperscript{\rm 5}, 
    Weihua Yue\textsuperscript{\rm 6}, 
    Yu Huang\textsuperscript{\rm 1}\thanks{Corresponding authors.}
}
\affiliations{
    \textsuperscript{\rm 1}Peking University, Beijing, China 
    \textsuperscript{\rm 2}School of Computer Science and Engineering, Beihang University, Beijing, China\\
     \textsuperscript{\rm 3}Beijing Jiaotong University, Beijing, China
    \textsuperscript{\rm 4}East China Normal University, Shanghai, China\\
    \textsuperscript{\rm 5}Xuanwu Hospital Capital Medical University, Beijing, China
    \textsuperscript{\rm 6}Peking University Sixth Hospital, Beijing, China\\
    joysw@stu.pku.edu.cn, zmliu@buaa.edu.cn, hy@pku.edu.cn

}

\begin{document}

\maketitle

\begin{abstract}
Online psychological counseling dialogue systems are trending, offering a convenient and accessible alternative to traditional in-person therapy. However, existing psychological counseling dialogue systems mainly focus on basic empathetic dialogue or QA with minimal professional knowledge and without goal guidance. In many real-world counseling scenarios, clients often seek multi-type help, such as diagnosis, consultation, therapy, console, and common questions, but existing dialogue systems struggle to combine different dialogue types naturally. In this paper, we identify this challenge as how to construct mixed-type dialogue systems for psychological counseling that enable clients to clarify their goals before proceeding with counseling. To mitigate the challenge, we collect a mixed-type counseling dialogues corpus termed STAMPsy\footnote{STAMPsy is publicly available at github.com/JOY-SWang/STAMPsy.}, covering five dialogue types, task-oriented dialogue for diagnosis, knowledge-grounded dialogue, conversational recommendation, empathetic dialogue, and question answering, over 5,000 conversations. Moreover, spatiotemporal-aware knowledge enables systems to have world awareness and has been proven to affect one's mental health. Therefore, we link dialogues in STAMPsy to spatiotemporal state and propose a spatiotemporal-aware mixed-type psychological counseling dataset. Additionally, we build baselines on STAMPsy and develop an iterative self-feedback psychological dialogue generation framework, named Self-STAMPsy. Results indicate that clarifying dialogue goals in advance and utilizing spatiotemporal states are effective. 

\end{abstract}

%

\section{Introduction}
One in eight people worldwide is living with mental health conditions, yet the growing demand for mental health care is facing significant challenges due to insufficient resources of existing consultants\cite{who2023mental}. The uneven distribution of face-to-face counseling resources and the high cost of therapy have exacerbated the situation of mental disorders\cite{ierardi2022effectiveness}. Thanks to the development of generative AI tools, such as Large Language Models(LLMs), online counseling is gradually becoming a good alternative to traditional face-to-face counseling \cite{OLAWADE2024100099, Message-Based_Psychotherapy}. 
 
\begin{figure*}[!ht]
\centerline{\includegraphics[width=0.95\linewidth]{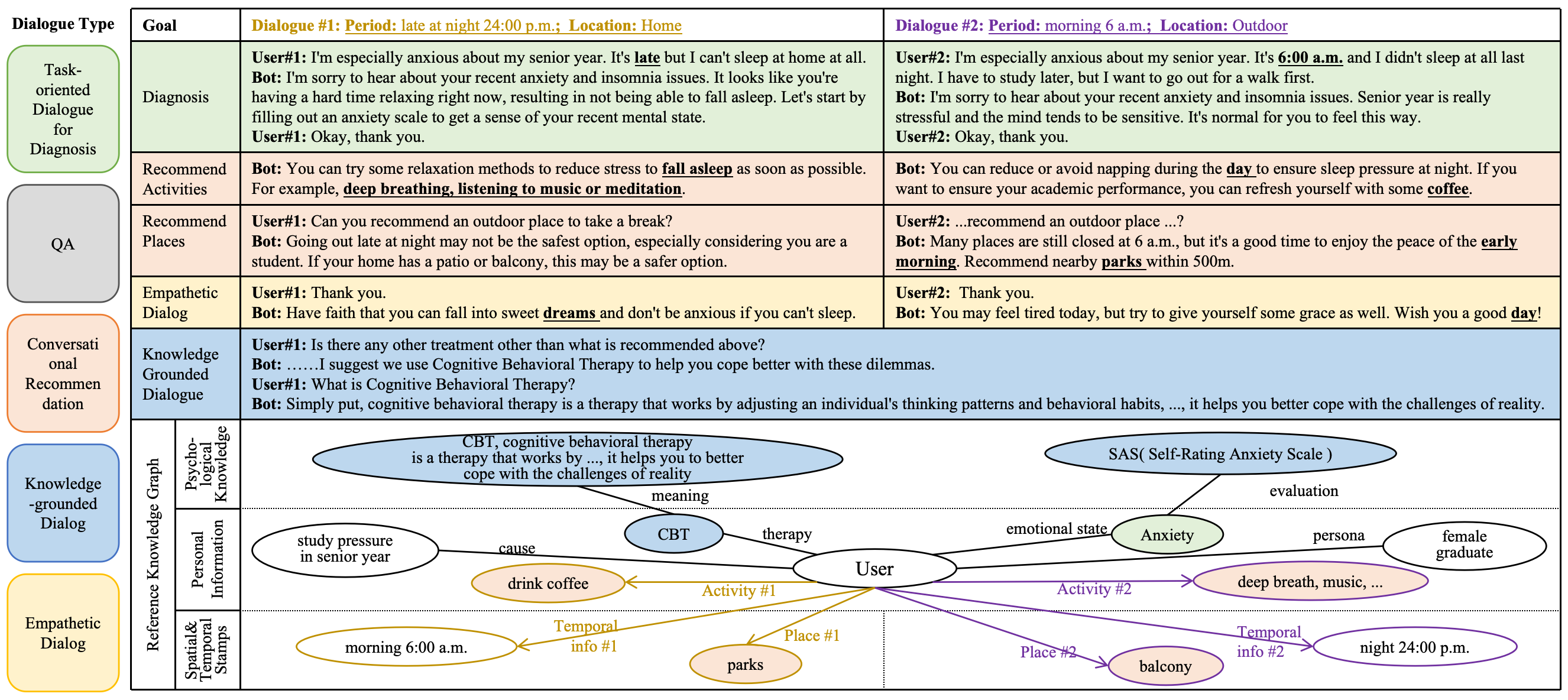}}
\caption{An example of STAMPsy with spatiotemporal state and reference knowledge.}
\label{img: dialoguesample}
\end{figure*}

Recent studies have leveraged LLM to aid clients in obtaining better online psychological counseling\cite{ke2024exploringfrontiersllmspsychological, wang2024promptengineeringhealthcaremethodologies}, 
falling into two categories: \ding{172}task-oriented dialogues or recent LLM-based simulations in psychology theory assume that clients have explicit goals (comfort-seeking, therapy querying, etc.)\cite{wang2024patientpsi, chen2023llmempoweredchatbotspsychiatristpatient}; \ding{173}non-task dialogue-modeling approaches with pre-trained models for single-turn QA or multi-turn empathic dialogue\cite{liu-etal-2021-towards, zheng-etal-2023-augesc}. Almost none of these studies naturally mix different types of dialogue(called \emph{mixed-type dialogues}). However, in real-world application scenarios, there are multiple dialogue types in psychological counseling. 
For example, as shown in Fig.\ref{img: dialoguesample}, the psychological counseling bots could proactively make QA or conversational recommendations after task-oriented dialogue for diagnosis to improve user experience.
However, to the best of our knowledge, no previous research has been done on this challenge.  
To mitigate this challenge, we present a novel task,  mixed-type dialogues for psychological counseling, as shown in Fig.\ref{img: dialoguesample}. 

Besides, beyond verbal conversations and body language, time and environment are two important nonverbal cues that affect counseling\cite{cormier2009interviewing} and mental health\cite{WooPostolache, timeaffectpsy}. Therefore, we delve into the spatiotemporal states for this task, leading the model to generate more world-aware answers. For recommending therapy, therapists generally give different advice for late-night insomnia and early morning awakening, relaxation exercises before bed, and avoiding long naps.

To facilitate the study of this task, we collate the first \textbf{S}patio\textbf{T}emporal-\textbf{A}ware \textbf{M}ixed-type dialogues dataset for \textbf{Psy}chological counseling(STAMPsy), stamped with spatiotemporal state and psychological helping skills, including five dialogue types: task-oriented dialogues for diagnosis, knowledge-grounded dialogue, conversational recommendation, empathetic dialogue, and QA, containing 5,006 dialogues and 61,832 utterances, with at least three dialogue types in each dialogue of STAMPsy. 
It should be noted that data processing is guided by psychological professionals. Specifically, we propose organizing the data according to the professional psychological theory of ``Case Conceptualization", including modules to gain the client's profile and personal information, spatiotemporal state, and goal, checked by professionals. 
As shown in Figure \ref{img: dialoguesample}, to help clarify the client's goal, the therapist proactively gives recommendations like scale testing and therapies, interspersed with emotional solace. Furthermore, the counselor's reply \textbf{changes with time and environment}. Although both suffer from insomnia, client \#1 hopes to fall asleep, while client \#2 hopes to stay alert during the day. Therefore, the counselor offers different suggestions, relaxation ways, and coffee respectively, which vividly demonstrates the effectiveness of spatiotemporal state on mixed-type dialogue.

To promote research on mixed-type dialogues for psychological counseling, we conduct bench-marking experiments on STAMPsy for the three sub-tasks:  
Helping Skills Selection, SpatialTemporal State Processing(STSP), and Adaptive Retrieval Augmented Generation. Moreover, inspired by \cite{Asai2023SelfRAGLT}, 
we propose a novel Iterative Self-Feedback framework termed \textbf{Self-STAMPsy} to self-reflect on the consultation process and optimize model, imitating the supervision report and fine-tuning pre-trained dialogue models for comparison with  \ding{172} different instruction texts as input and  \ding{173} adaptive retrieval-augmented generation from knowledge graph. Experimental results also demonstrate that utilizing the above four modules improves the answers.

Our contributions are summarized as follows:
\begin{itemize}
    \item To the best of our knowledge, we are the first to identify the new challenge that there are multiple dialogue types in psychological counseling for online clients. 
    \item To mitigate this challenge, we propose the first Chinese mixed-type dialogue dataset stamped with the spatiotemporal state and goals for psychological counseling (STAMPsy), where the spatiotemporal state of counseling is stamped to better connect the psychological dialog system with reality and explore clients' potential emotional responses to their surroundings.
    \item We build baselines on STAMPsy and propose a novel iterative self-feedback framework Self-STAMPsy simulating the supervision process. Extensive experiments demonstrate that helping skill, spatiotemporal state and self-feedback retrieval can improve performance.
\end{itemize}

\begin{table*}[ht]
\resizebox{17.65cm}{!}{ 
\centering
\begin{tabular}{lcccl}
\hline
Datasets                      &      S.T.-aware   & Mixed-type & Psy. & Dialogue Types                                             \\ \hline
ESConv\cite{liu-etal-2021-towards}                               & \textcolor{red}{$\times$}   &  \textcolor{red}{$\times$}  & \textcolor{green}{\checkmark}                        & Emotional response generation                              \\

PsyQA\cite{sun-etal-2021-psyqa}                   &  \textcolor{red}{$\times$}                &  \textcolor{red}{$\times$}   & \textcolor{green}{\checkmark}                       & single-turn conversations for mental health support        \\
AugESC\cite{zheng-etal-2023-augesc}                               &  \textcolor{red}{$\times$}   &  \textcolor{red}{$\times$}    & \textcolor{green}{\checkmark}                      & Large emotional response generation                        \\
D4\cite{Yao2022D4}         &  \textcolor{red}{$\times$}  & \textcolor{red}{$\times$}          & \textcolor{green}{\checkmark}           & Chitchat, task-oriented dialogue for depression   \\
CPsyCoun\cite{zhang2024cpsycoun} &  \textcolor{red}{$\times$}   &  \textcolor{red}{$\times$}    & \textcolor{green}{\checkmark}                      &  Multi-turn dialogues with different consultation topics \\                    
\hline

BlendedSkillTalk\cite{smith-etal-2020-put}  &  \textcolor{red}{$\times$}  & \textcolor{green}{\checkmark}          &  \textcolor{red}{$\times$}                   & Know., empathetic dialogue, chitchat                        \\
DuRecDial 2.0\cite{liu-etal-2021-durecdial}  &  \textcolor{red}{$\times$}   & \textcolor{green}{\checkmark} &  \textcolor{red}{$\times$}  & Rec., chitchat, QA, task-oriented dialogue           \\
                         
DuClarifyDial\cite{liu-etal-2022-go}    &  \textcolor{red}{$\times$}   & \textcolor{green}{\checkmark}&  \textcolor{red}{$\times$} & Rec., know. chitchat, QA, task-oriented dialogue     \\
MidMed\cite{shi-etal-2023-midmed}         &  \textcolor{red}{$\times$}  & \textcolor{green}{\checkmark}          &  \textcolor{red}{$\times$}             & Rec., empathetic dialogue, know., QA, diagnosis-oriented dialogue                 \\
\hline
DuSinc\cite{zhou2022link}           & \textcolor{green}{\checkmark}                      &  \textcolor{red}{$\times$}           &  \textcolor{red}{$\times$}                        & Service information augmented dialogue \\ 

\hline
STAMPsy (Ours)                      & \textcolor{green}{\checkmark}          & \textcolor{green}{\checkmark}             & \textcolor{green}{\checkmark}                    & Rec., chitchat, know., QA, diagnosis-oriented dialogue     \\ \hline
\end{tabular}
}
\caption{Comparison of STAMPsy with other datasets. ``Psy.", ``S.T.-aware", ``know.", and ``rec." stand for psychological counseling dialogue, spatiotemporal-aware knowledge, knowledge-grounded dialogue,
and recommendation, respectively.}
\label{tab:dataCompar}
\end{table*}

\section{Related Work}
\subsection{Applications of LLMs in Psychology}
Recently, large language models(LLMs) with professional expertise and much better inferential capability have the potential to converse like a real person\cite{wang2024learningbreakknowledgeenhancedreasoning,tao2024evit}, leveraging their capabilities for tasks like reasoning and interaction.
Especially in psychology, LLMs are proven capable of replacing human participants in experiments\cite{apa1}, like emulating human social dynamics\cite{park2023generative, wang2024userbehaviorsimulationlarge}. Therefore, we make LLMs serve as therapists and clients to generate data similar to counseling dialogues.  
However, previous research has mainly focused on simply establishing an identity for LLMs, with limited relative information and without considering complex scenarios in real-world counseling\cite{cho2023evaluating}. We include the spatiotemporal state during counseling and the goal sequence of the client-counselor dialogue, which influences the client's emotional state, and is crucial for assisting LLMs in better solving clients' problems. 

\subsection{Datasets for Mental Health Support}
Research on mental health support typically relies on high-quality psychologically pertinent datasets. For example, ESConv \cite{liu-etal-2021-towards} suggests using programs and dialogue techniques to offer emotional support through exploration, comfort, and action. AugESC \cite{zheng-etal-2023-augesc} expands the ESConv dataset with LLMs. 
CPsyCoun\cite{zhang2024cpsycoun} contains multi-turn dialogues for various consultation topics. PsyQA\cite{sun-etal-2021-psyqa} analyzes the organized psychological strategy sequence but gets single-turn QA. Although these datasets enhance the performance of LLMs in psychology, most of them mix all types of dialogues, leaving the model unable to give more targeted answers. Table \ref{tab:dataCompar} lists high-quality psychological and mixed-type dialogue datasets. However, they primarily focus on pure psychological counseling or cannot capture and perceive other mixed information that might be used in the counseling process. Faced with the absence of a dataset for mixed psychological counseling conversations, we proposed STAMPsy, added mixed dialogue goals to conversations, and conducted in-depth research on the positive impact of the spatiotemporal state on counseling conversations. This could be the initial step towards a spatiotemporal perception of mixed dialogue similar to real-world psychological counseling.


\subsection{Mixed-type Dialogue Systems}
Recent research on mixed-type dialogue has witnessed a marked upsurge, especially in open-source datasets. Some researchers train a unified, comprehensive conversation model\cite{madotto2020adapterbot,roller2020recipes}, which integrates multiple dialogue skills into a singular framework by combining various single-skill conversation
datasets, such as persona-chat and task-oriented dialogue. Others focus on mixed-type dialogue datasets and models, designed to aggregate diverse dialogue skills to meet specific needs, like recommending music and places, but unable to solve psychological problems\cite{liu-etal-2020-towards-conversational, liu-etal-2021-towards}. Compared with them, we compiled the first mixed-type spatiotemporal dialogue dataset for psychology, which examines the effectiveness of integrating counselor helping skills and fusing spatiotemporal state in online psychological counseling.

\section{Dataset Collection} 
\begin{figure*}[!ht]
\centerline{\includegraphics[width=0.99\linewidth]{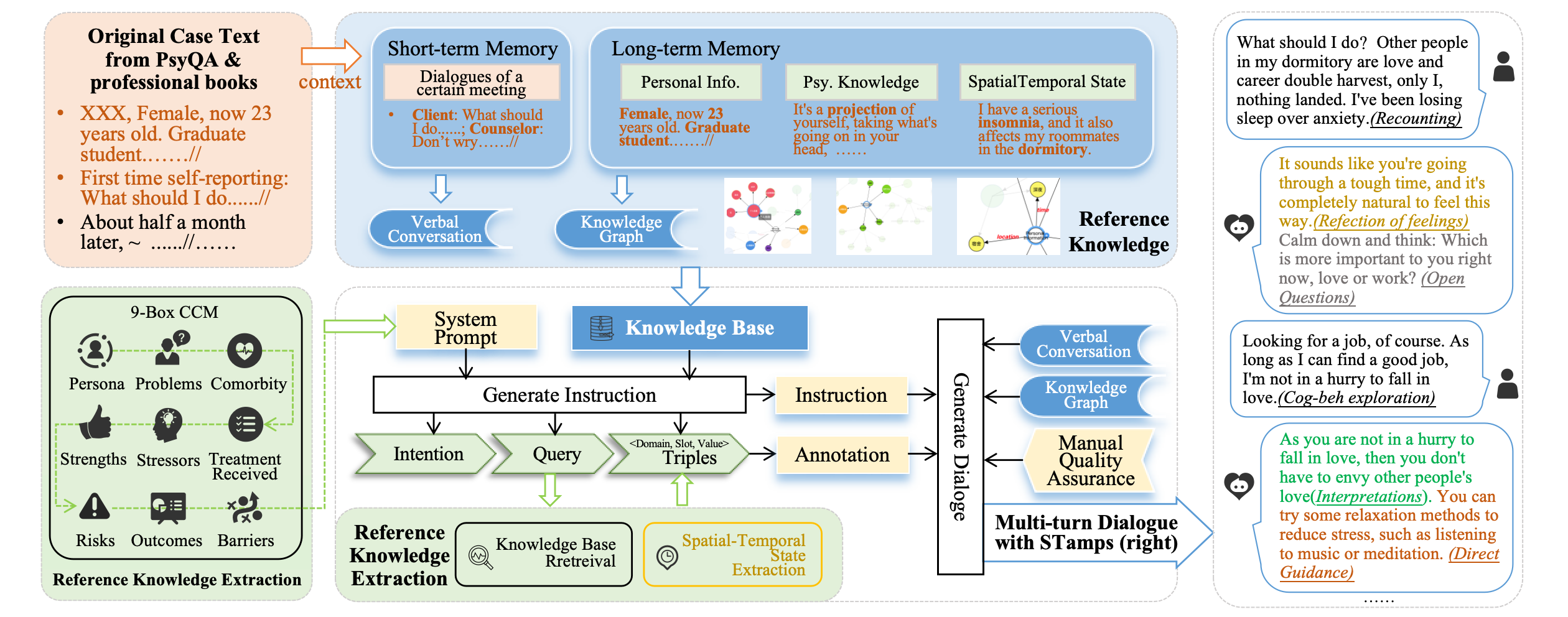}}
\caption{The collection workflow of STAMPsy. We extract reference knowledge from context under 9-Box CCM and then gain multi-turn dialogues with a multipart instruction. All the dialogues are annotated and revised by psychological experts.}
    \label{fig:Agent4Psy}
\end{figure*}

This paper aims to construct a spatiotemporal-aware mixed-type dialogue dataset for online psychological counseling. To more authentically reflect the patterns of real-world psychological counseling, we devise multiple modules to mimic practical counseling case report patterns. 
The construction flowchart is presented in Figure \ref{fig:Agent4Psy}.


\subsubsection{Data Source}
Given privacy concerns, it is difficult to obtain real-world consultation records. Therefore, we first collected a 2.6GB knowledge base with a large number of psychologically oriented books, scales, past counselor exam papers, and other materials. Then, we 
analyzed public cases and in-depth descriptions from popular counseling books\cite{kamaPsychologicalCases, cormier2009interviewing, hill2009helping}. 
To better fit the online context, we also selected conversations from PsyQA\cite{sun-etal-2021-psyqa}, a Chinese dataset of psychological health support with the Helping Skills Theory. To ensure the controllability of the content generated by the agent system, the topics are in the ideal topic list for the laboratory activity(see Appendix B). 
After collection, we filter private information, divide long cases into several meetings manually, and ultimately collect 4,000 single texts as the origin input.



\subsection{Reference Knowledge Extraction} 
Psychological counseling tends to be a long, multi-turn dialogue, and response quality hinges
on communication coherence and personal information
consistency. Inspired by \cite{park2023generative}, we use short-term memory to
update the context. For long-term memory, continuous interactions, personal information, psychological knowledge, and spatiotemporal stamps help maintain persona consistency.
We utilize LLMs serving as structured psychological assistants to capture information from original case text and collate them into triples in the form of $[Domain|Slot|Value]$. 
``Domain" includes ``Personal Information", ``Spatial-Temporal Information" and ``Psychological Knowledge". 

\paragraph{Personal Information} Directed by 9-BOX Case Conceptualization Model\cite{meichenbaum2009psycho}, we collate ``Personal Information" with nine slots step by step, including \ding{172}Personal Profile and Background, \ding{173}Problem Presentation, \ding{174}Comorbidity, \ding{175}Stressors, \ding{176}Treatments Received with their efficacy, adherence, and satisfaction, \ding{177}Strengths, \ding{178}Summary Risk and Protective Factors, \ding{179}Outcomes and \ding{180}possible Barriers. More details are shown in Appendix A.

\paragraph{Spatial-Temporal Information\label{sec:STIE}} The increasingly popular online text psychological support enables counseling to take place anytime, anywhere\cite{LIU2021104367}. However, previous online counseling dialogue systems cannot comprehend complex scenarios with limited information. 
Therefore, we focus on the spatiotemporal state to better connect the reality.
Moreover, spatial effects, environment, and time are three important nonverbal behaviors in communication because people have emotional responses to their surroundings\cite{cormier2009interviewing}. 
Accordingly, for each round of conversation, we extract and label non-private spatiotemporal state ${ST}_i$ such as time, location, or weather based on regular expressions, and conclude their influence on mental health to spatiotemporal stamps. More details are shown in Appendix C.

\paragraph{Psychological Knowledge} The collection of knowledge from the dialogues and the original knowledge base is converted to triples.
Inspired by the construction and application of the temporal knowledge graphs
(TKGs)\cite{trivedi2017know, trivedi2019dyrep}, where the fact extends from a triple $(s, p, o)$ (respectively representing the subject, predicate as a relation type, and object, and saved as memory after a manual check by psychological experts) to a quadruple with a timestamp $t$.  We collect an event and redefine it as a spatiotemporal state stamped edge represented as a quadruple, $[Domain|Slot|Value|Stamp]$. For ``Relaxing Method Recommendation" slot, the value contains ``drink coffee" with the timestamp ``morning" but ``read a book" to ``night".

Based on reference knowledge quadruples, we conclude a multipart system prompt shaped by instructions based on the CCM and expand single-turn text to multi-turn dialogues. What's more, we invite clinical psychologists to check the consistency and correctness of the dialogues.

\subsection{Dataset Annotation}
Dataset annotation involves labeling dialogue goals and exploring knowledge graphs, done by 8 psychology experts. The goals are split into counselor's helping skill\cite{hill2009helping} and client's behaviors\cite{Hill1992}(right of Fig. \ref{fig:Agent4Psy} in parentheses), with detailed guidance(see Appendix E) and knowledge triples provided to assist annotation. To ensure annotators understand the process, a trial annotation precedes the formal one.
\subsubsection{Target Helping Skills Sequence}
 During trial annotation, we extensively analyze the generated conversations above and find that the goals sequence patterns in the dialogue are similar to Counselor Helping Skills commonly used in psychology\cite{bickmore2011reusable, hill2009helping, ActivitySelf-EfficacyScales}. Thus, we assume that
each counseling can be realized through an organized helping skill sequence, which may reveal the common
the layout of high-quality mixed-type conversation in psychological
counseling, simplifying the process of data annotation. 

\begin{figure}[htbp]
\centerline{\includegraphics[width=0.95\linewidth]{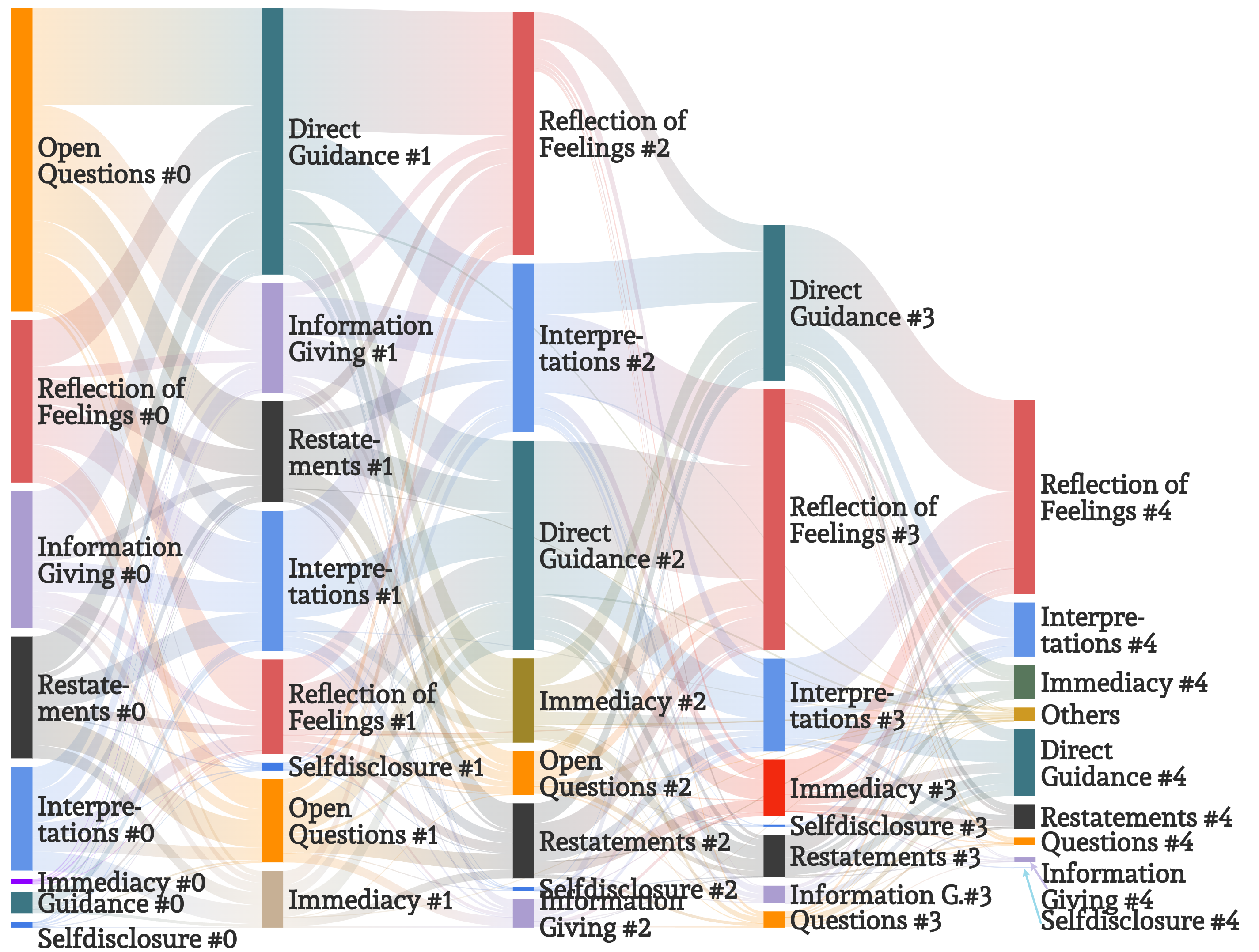}}
\caption{Sequence visualization of the common dialogue goal
flow patterns within the first 5 counselor-helping skills.}
\label{topics}
\end{figure}

Inspired by MultiWOZ\cite{budzianowski-etal-2018-multiwoz}, for each dialogue session, we provide a targeted dialogue helping skill sequence assigned by at least three of eight helping skills\cite{Hill1986}, which are ``Immediacy"(Imme.), ``Interpretations"(Inptn.), ``Self-disclosures", 
``Open questions", ``Feeling Reflection"(Feel.), ``Restatements"(Rest.), ``Information giving"(Info.), ``Direct guidance"(Guid.) and ``Others". Direct guidance can be categorized as recommendations of places, relaxing ways, lifestyles, therapies, and music. Specifically, thanks to spatial stamps, the ``Direct guidance: Recommended Place" label aims to guide users to a private place while counseling outdoors or in a crowded environment. These helping skills can also be mapped to the five types of conversation(see Tab. \ref{tab:activitiesStat}). The whole dialogue
sequences with the first five distinct helping skills are shown in Figure \ref{topics}. Based on the analysis above, we annotate each dialogue with Client Behaviors and Counsellor Helping Skills according to Hill Counselor Verbal Response Category System (a detailed system description can be found in Appendix D).

\begin{table}[ht]
\centering
\resizebox{0.48\textwidth}{!}{
    \begin{tabular}{llccc}
    \hline
    \multicolumn{2}{l}{\textbf{Goal Type}} & \textbf{\# Num} & \textbf{Avg.Len.} & \textbf{Helping Skill}      \\ \hline
    \multicolumn{2}{l}{Diagnosis} & 3745 & 85.71          & Imme., etc.  \\ 
    \multicolumn{2}{l}{QA}        & 4500 & 89.82          & Intpn., etc. \\ 
    \multicolumn{2}{l}{Knowledge-grounded dialog}     & 4857 & 67.06          & Info., etc.  \\ 
    \multirow{6}{*}{\begin{sideways}Recommend \end{sideways}}    & Place   & 89              & 75.15              &  \\
           & Relaxation          & 1128 & 82.72          &              \\
           & Lifestyle       & 2210 & \textbf{97.48} & Direct             \\
           & Therapy         & 1777 & 91.69          & Guidance             \\
           & Music           & 137  & 74.46          &              \\
           & Total           & 5330 & 89.3           &              \\ 
    \multicolumn{2}{l}{Empathetic dialog}   & \textbf{7229}   & 66.90              & Feel., etc.      \\ \hline
    \end{tabular}
}
\caption{The number and the average length of goal types and counselor helping skills.}
\label{tab:activitiesStat}
\end{table}

\subsubsection{Knowledge Graph Construction}
We incorporate knowledge graphs, archiving extensive knowledge in the form of triples, into psychological counseling to provide more accurate interactive questions and answers\cite{yang-etal-2020-graphdialog}. After experts recheck the correctness and completeness of triples, annotators relate entities about disorder, symptom, therapy, etc., and relationships including disorder-symptom relation, disorder-therapy relation, etc, to construct a psychological knowledge graph $K$. For each dialogue, we add a subset including persona and spatiotemporal stamps.

\subsection{Data Quality Audit and Analysis}
All annotated samples are thoroughly reviewed to identify low-quality dialogues and remove contentious illusions or hazardous information. Substandard samples are re-annotated. 
Following \cite{liu-etal-2020-towards-conversational}, we employ two senior psychologists for data quality evaluation on 1,000 randomly sampled dialogues. Specifically, they assign ``1" for dialogues following annotation guidance, and ``0" for the rest. The final average evaluation score of our dataset is ``0.91", with a Kappa value\cite{tinsley1975inter} of ``0.84", 
indicating that the dialogues are of high quality. 
Finally, we propose a Chinese SpatioTemporal-Aware Mixed-type multi-turn dialogues dataset for Psychological counseling. As shown in the bottom right corner of Fig.\ref{fig:Agent4Psy}, each meeting consists of multiple rounds of dialogues with at least three distinct Helping Skills. 

\begin{figure*}[!htp]
\centerline{\includegraphics[width=0.99\linewidth]{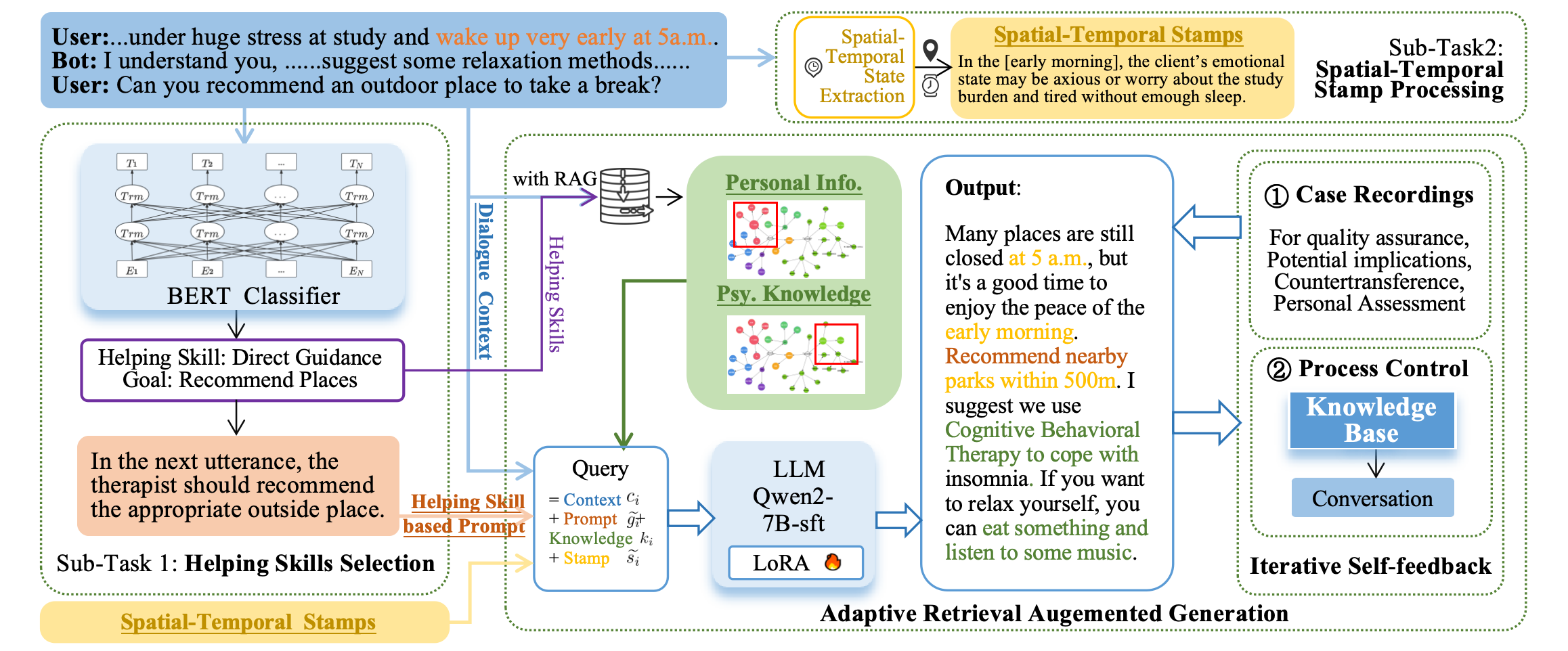}}
\caption{The framework of the proposed Self-STAMPsy. A detailed prompt template is open-sourced.}
\label{modelframework}
\end{figure*}

Table \ref{tab:activitiesStat} shows the number and the average length of goal types and counselor helping skills. 
As we can see, the empathetic dialogue is the most common activity because emotional solace such as encouragement, validation, and consolation are of great significance in psychological counseling\cite{lee2007validation}. In contrast, diagnosis is relatively rare, where external knowledge and backgrounds are extra required. We also noted that the average lengths of empathetic dialogue are remarkably shorter than other activities, where emotional solace that helps clients clarify or explore their thoughts or feelings is feasible with fewer words. What's more, we found that Diagnosis, QA, Know., and Rec. These four types of activities are more inclined to contain psychological knowledge, which means helping skills like Immediacy, Interpretations, Infomation Giving, and Direct Guidance tend to involve external knowledge, and that answer can be enhanced with RAG. Therefore, we label these four types of helping skills as $ HS_{wk}$, which means helping skills with knowledge. In the RAG modeling process, we first predict the category of the assistive technology (as depicted by the purple arrows in the figure\ref{modelframework}), and then further select to perform adaptive RAG.

\begin{table}[ht]
\centering
\resizebox{7.8cm}{!}{
\begin{tabular}{lc}
\hline
\textbf{Criteria Statistics}      & \textbf{Statistics} \\ \hline
\# of dialogues                      & 5,006                \\
\# of utterances of User/Bot             & 24,762/25,661   \\
\# of counselor helping skills             & 8   \\ \hline
Avg. \# of tokens per User          & 36.7                \\
Avg. \# of tokens per Bot       & 78.9                \\ \hline
Avg. \# of goals per dialogue          & 16.51         \\
Max. \# of goals per dialogue          & 34         \\
Min. \# of goals per dialogue          & 6         \\

\begin{tabular}[c]{@{}l@{}}Avg. \# of distinct activities per dialog\end{tabular}  & 3.74                    \\ \hline
\end{tabular}
}
\caption{\label{tab:datasetStat} Statistics of our dataset. }
\end{table}

Additionally, the detailed statistics of our dataset are listed in Table \ref{tab:datasetStat}. The range of goals per dialogue (from 6 to 34) shows variability in the complexity and length of the dialogues. The average number of distinct goals per dialogue (3.74) suggests that there are at least three different types of conversation goals for each meeting.



\section{Methodology}
\subsection{Overview}
Recently, the methodologies for different types of dialogues have gradually shifted to generative and end-to-end modeling. Following these trends, we propose a pre-trained mixed-type dialogue model \textbf{Self-STAMPsy} based
on Qwen2\cite{bai2023qwen}. During training, a dialogue with a sequence of utterances between a client and a counselor is given. Then the dialogue is processed into a set of samples $\{(c_i, t_i)\}$, where $t_i$ is $i_{th}$ target counselor response,
$c_i$ is the concatenation of all former utterances before $t_i$, and $D$ is the train set. Dialogue generation is formulated as an end-to-end dialogue generation(E2EDG) task, which aims to
generate $t_i$ conditioned on $c_i$ as a query into LLMs.

Accordingly, Self-STAMPsy has four modules, helping skills selection, spatiotemporal state extraction, adaptive retrieval augmented generation, and iterative self-feedback. Helping skills selection and spatiotemporal state extraction aim to obtain dialogue goals $\widetilde{g_i}$ and
spatiotemporal stamps $\widetilde{{ST}_i}$, respectively. 
Moreover, by integrating a novel iterative self-feedback mechanism, we get the final response $t_i$.
The framework of Self-STAMPsy is plotted in Figure \ref{modelframework}. 

\subsection{Helping Skills Selection}
The Helping Skills Selection sub-task takes dialogue context and current dialogue goal sequence as input, then outputs predicted counselor helping skill. Formally,
this task is regarded as a multi-class classification
problem. Specifically, this module is divided into
two baseline models, the dialogue goal prediction with or without context. 
The input of the prediction module is a dialogue context $c_i$. With context $c_i$, it outputs the counselor helping skill $h_i $  and predicts recommended goal $g_i$. For the baseline model without
contextual information, The classification process is formulated,
$$p_i = f(c_i),$$
where $f$ is the classification function BERT\cite{devlin-etal-2019-bert} and $p_i\in {\left | \mathcal{R}  \right | }^{\left | \mathcal{G}  \right |}$  is the predicted
probability value, $\mathcal{G} $ is the predefined category set.
The dialogue activity $g_i$ is selected as the predicted dialogue activity if the value of the dimension is the highest probability value in $p_i$. For the model with context, we input multiple consecutive sentences $\mathbf{c_1, c_2, c_3, ...}$ to BERT in the form
of [CLS]$\mathbf{c_1}$[SEP][CLS]$\mathbf{c_2}$[SEP][CLS]$\mathbf{c_3...}$ and compute the mean loss of [CLS] located at the
beginning of each sentence. After Helping Skills Selection, to promote the effectiveness, we convert dialogue goal $g_i$
into natural language with predefined
templates, represented as $\widetilde{g_i}$. If the predicted goal is ``Recommend Therapy", the converted instruction will be ``The therapist will then design a therapy".

\subsection{SpatioTemporal Stamp Processing}
SpatioTemporal State Processing(STSP) contains two steps, spatiotemporal state extraction and stamp generation. The state ${ST}_i$ can be obtained from the Memory Module. 
Subsequently, we utilize the position embeddings of $c_i$ and ${ST}_i$ as inputs and derive emotional state $\widetilde{{ST}_i}$ by employing the mechanism previously analyzed for the goal. During training, we minimize the negative log-likelihood(NLL) loss:
\begin{equation}
\notag
\begin{aligned}
\mathcal{L}_{NLL_{\widetilde{{ST}_i}}} & =-\mathbb{E} \log p(\widetilde{{ST}_i} \mid {ST}_i, c_i) \\
& =-\mathbb{E} \sum_{i=1}^{n} \log p\left(\widetilde{{ST}_i} \mid {ST}_i, c_i, \widetilde{{ST}_i}_{<i}\right)
\end{aligned}
\end{equation}

where $\widetilde{{ST}_i}<i$ denotes previously generated emotional state in
query $Q$. Then based on the impact of time and place on one's emotions, we convert $S$ into emotional state $\widetilde{{ST}_i}$. 

\subsection{Adaptive Retrieval Augmented Generation}
RAG methods enhance the input of LLMs with pertinent retrieved passages, thereby minimizing factual inaccuracies in knowledge-intensive tasks\cite{realm2020, NEURIPS2020_6b493230}. 
Graph-RAG\cite{edge2024localglobalgraphrag} has claimed the use of graphs in connection with LLMs and RAG is powerful. Therefore, we combine knowledge graph $K$ with LangChain, a RAG tool, and get augmented response $t_i$.  


\paragraph{Iterative Self-feedback}
To ensure interaction quality and dialogue consistency, we add the Iterative Self-feedback mechanism. We design a multipart prompt with
two main modules:  \ding{172}\textbf{Case Recordings} for quality assurance. In counseling, the counselor is required to reflect on their client between every two sessions, trying to get the causes of their problems, the underlying themes behind the problems, and the interventions that apply to helping the client. Refering to Hill\cite{hill2009helping}, we designed the model to self-reflect each turn, including six parts: explicit content, implicit content, barriers to defense and change, distortions, countertransference, and personal assessment. 
\ding{173}\textbf{Process Control} utilizes historical interactions to determine the optimal timing to end dialogues. If the conversation isn't over $(i<n)$, the output $t_i$ will be used for iterative model optimization.

\begin{table}[!ht]
\centering
\resizebox{6.7cm}{!}{
\begin{tabular}{cccc}
\hline \textbf{Helping Skill} & \textbf{Prec.} & \textbf{Recall} & \textbf{F1} \\
\hline \multirow{2}{*}{ Immediacy } & \textbf{64.84}    &   \textbf{26.92}    &   \textbf{38.05} \\&
59.14    &   22.58    &   28.09 \\
\hline \multirow{2}{*}{ Interpretations } & \textbf{ 77.92}    &   \textbf{87.57}    &  \textbf{ 82.46} \\&
71.68    &   81.71    &   75.79\\
\hline \multirow{2}{*}{ Open questions } & 93.97    &   93.80    &   94.38 \\&
\textbf{96.80}    &   \textbf{97.39 }   &  \textbf{ 96.08}  \\
\hline \multirow{2}{*}{ Feeling reflection} & 84.85    &   80.31    &   82.52 \\&
\textbf{87.31 }   &   \textbf{82.81 }   &  \textbf{ 85.00}  \\
\hline \multirow{2}{*}{Restatements } & \textbf{ 70.03}    &   31.49    &   43.44 \\&
68.95    &  \textbf{ 41.87  }  & \textbf{  48.96 } \\
\hline \multirow{2}{*}{ Information giving } & 58.44    &   48.75    &   53.16 \\&
\textbf{69.68 }   &   \textbf{60.66 }   &   \textbf{64.42  }\\
\hline \multirow{2}{*}{ Direct guidance } & \textbf{83.34}    &   78.48    &   80.84 \\&
82.80    &   \textbf{87.42}    &   \textbf{85.05 }  \\
\hline \multirow{2}{*}{ Weighted avg. } & 78.89    &   70.47    &   73.13 \\&
\textbf{79.40 }   &   \textbf{74.73}    &   \textbf{75.97}  \\
\hline
\end{tabular}
}
\caption{Sub-Task1: Helping Skills Selection. The BERT strategy classification results for each helping skill. We compare the performance between the models without(the upper row) or with(the lower row) context.}
\label{tab:subtask2withACTs}
\end{table}

\section{Experiments}

\subsection{Baselines}
Following PsyCoun\cite{hu2024psycollmenhancingllmpsychological}, we carefully select five fine-tuned robust LLMs, GPT4, and two Chinese psychological models as baselines to explore and evaluate the performance of various models.
To optimize network performance, we perform LORA\cite{hu2022lora} and use LLaMa-Factory\footnote{https://github.com/hiyouga/LLaMA-Factory.} based on STAMPsy and under the same parameters. 

\subsection{Experimental Setting}
For the BERT classifier, we use a mini-batch size of 128 and the Adam optimizer with default parameters (fixed learning rate 0.001, $\beta_1 = 0.9, \beta_2 = 0.999, \epsilon = 1 \times e^{-8}$) \cite{kingma2014adam}. For finetuning, we add a dense output layer on top of the model with a cross-entropy loss function. 
Our experiments are conducted on the workstation with Linux 6.1.0-10-amd64, four NVIDIA A800 GPUs, and Debian 10.

\begin{figure*}[htbp]
\centerline{\includegraphics[width=0.95\linewidth]{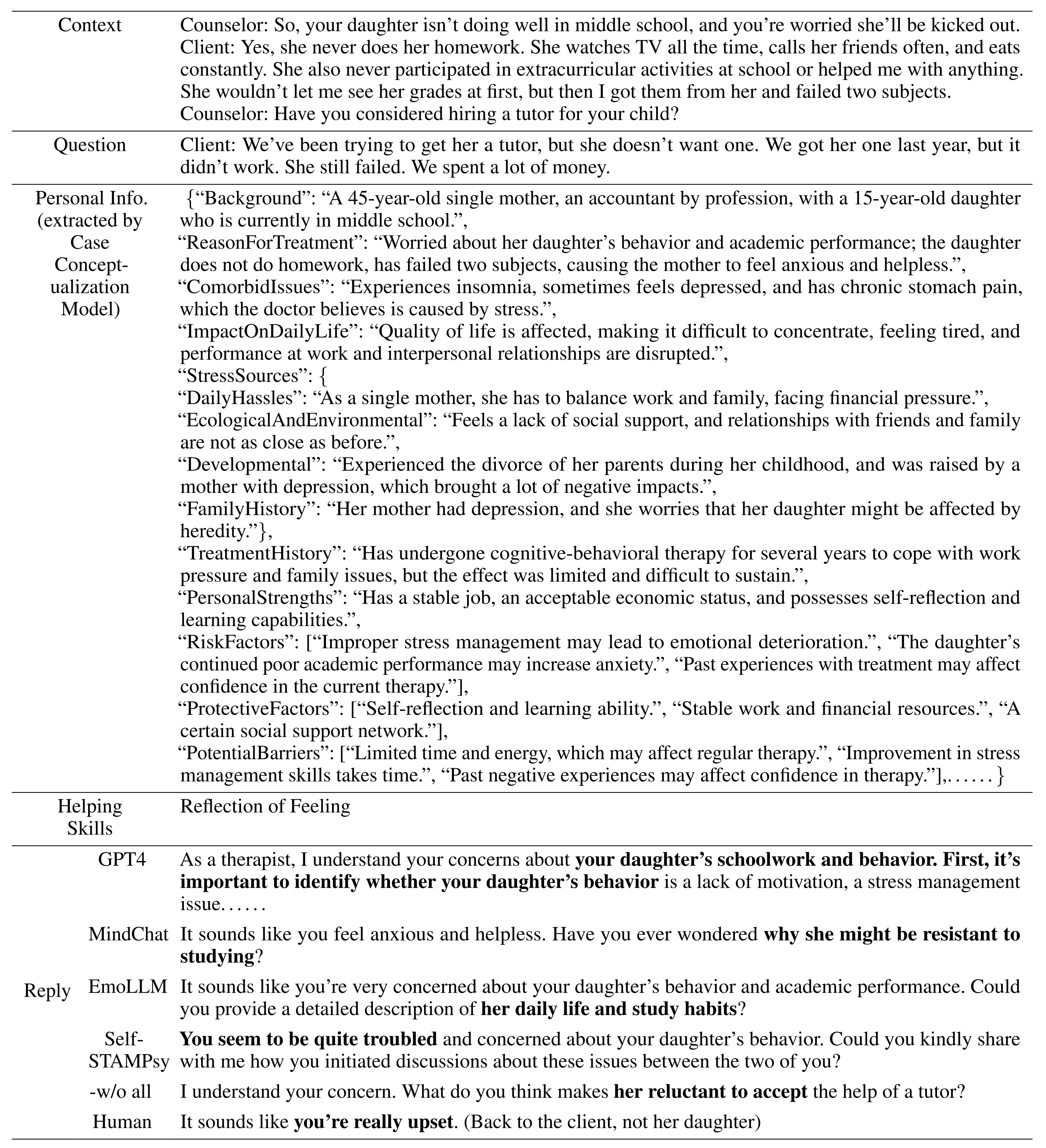}}
\caption{A case of generated answers and the golden answer. It can be observed that without the guidance of counseling techniques, other models tend to explore how to improve children's learning, which appears more like a communication about education rather than addressing the emotional concerns of the visitor, which is different from golden answer in the professional books. Subsequent conversations should revert to focusing on the client's personal issues, so it is better to come into play the ``emotional reflection" counseling technique. In our model, the fine-tuned BERT model first correctly categorizes the helping skill. Based on the goal of ``Reflection of Feeling", Self-STAMPsy can achieve better response effects, returning to the client's own concerns.}
\label{caseStudy}
\end{figure*}

\begin{figure*}[htbp]
\centerline{\includegraphics[width=0.95\linewidth]{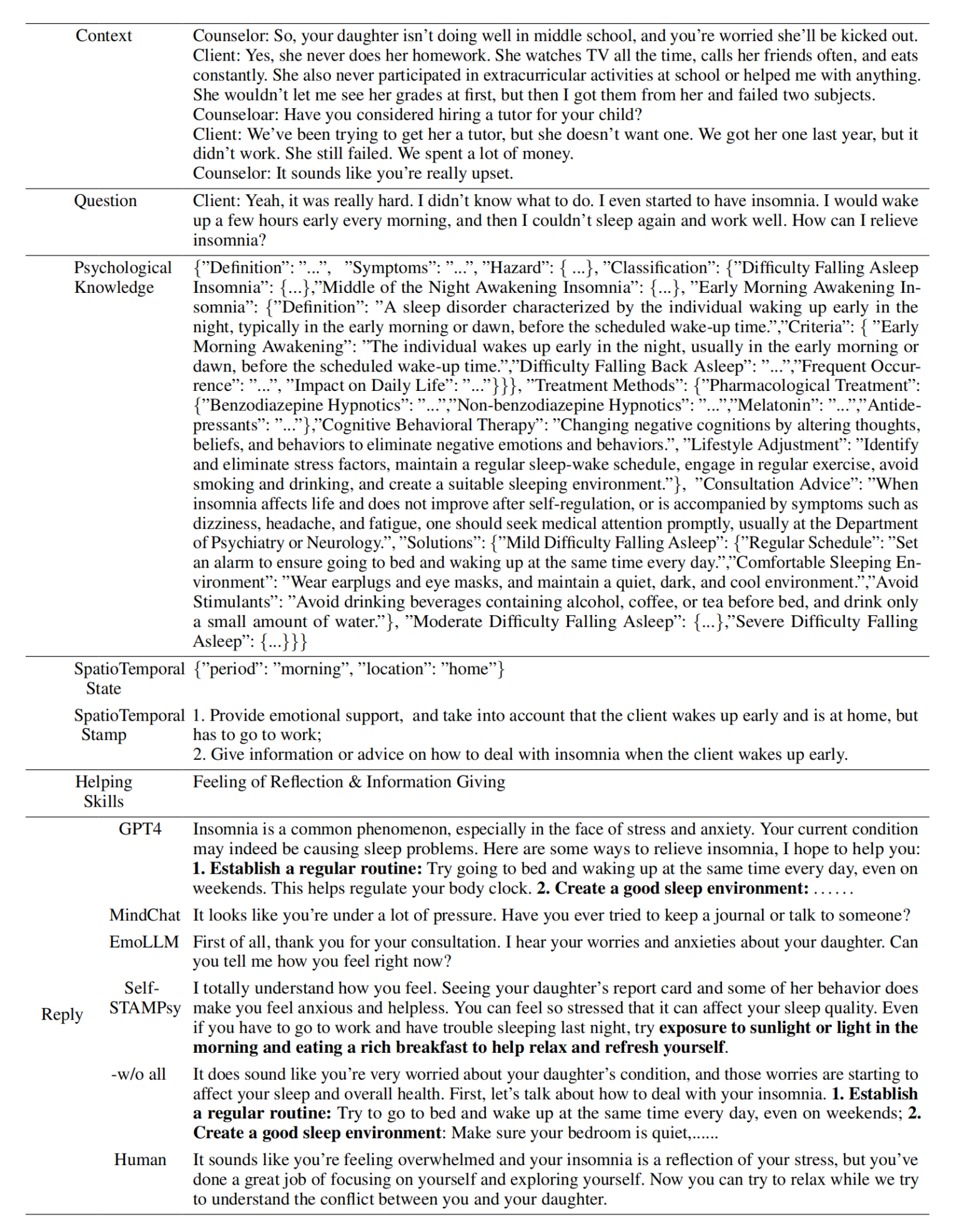}}
\caption{A case of generated answers and the golden answer, focusing on the spatiotemporal state.}
\label{caseStudy4ST}
\end{figure*}

\subsection{Evaluation}
For Helping Skills Selection, we calculate the precision, recall, and f1-score in the test set as a classification task. In addition, to assess whether the model truly understands the helping skills, we devise a benchmark, called GHSC, following Hill Counselor Verbal Response
Category System\cite{Hill1992} and evaluate all models on GHSC.  
For STSP, we calculate the accuracy.
As mentioned above, the problem of response generation is formulated as an E2EDG task. The input to the generation model is the concatenation of the dialogue context $c_i$, the
predicted dialogue helping skill prompt $\widetilde{g_i}$
, the reference knowledge ${k_i}$ and the spatiotemporal stamp $\widetilde{{ST}_i}$. The output is the counselor’s response $t_i$.
To evaluate the performance of E2EDG, we utilize both automatic metrics and manual metrics.
For automatic evaluation, we use several classical metrics, including BLEU-$n$ \cite{Papineni2002BleuAM}, ROUGE\cite{Lin2004ROUGEAP}, and BertSim\cite{kexuefm7427}.
For manual evaluation, we asked four psychotherapists to evaluate sessions from each end-to-end conversation with consensus judgment, encompassing five feature indicators: Relevance(Rel.), Informativeness(Info.), Human-likeness(Human.), Helpfulness(Help.), Empathy(Emp.). The specific description of the indicators and their scoring criteria are stated in Appendix I.


\subsection{Results}
To ensure that the retrieved knowledge in sub-task 2 ``SpatioTemporal Stamp Processing" is correct, we carried out Dialogue State Tracking. Predicting ``Slot" in triples in the knowledge graph can achieve a high accuracy rate, as shown in Table \ref{subtask12}, which indicates
the effectiveness of utilizing LLMs as the backbone.

\begin{table}[ht]
\centering
\begin{tabular}{l|ccccc}
\hline
Methods      & Slot   & Value & $ST_i$  \\ \hline
GPT4          & 99.16 & 58.56  & 72.50   \\
ChatGLM3      & 98.86 & 45.96  & 63.50   \\
Baichuan2    & 99.09 & 48.51  & 68.02     \\
BLOOMZ        & 98.07 & 41.93  & 61.13     \\
LLaMa3-zh    & 98.64 & 41.95  & 55.37     \\ \hline
Self-STAMPsy    & \textbf{99.21} & \textbf{59.94}  & \textbf{76.08}    \\
w/o Helping Skill & 99.09 & 46.22  & 72.82    \\ \hline
\end{tabular}
\caption{\label{subtask12}
Sub-Task2: Detailed SpatioTemporal Stamp Processing Results on Self-STAMPsy, GPT4 and another 4 fine-tuned models. We measure the ``Slot" accuracy and Rouge-L for ``Value". The $ST_i$ column shows the accuracy of the spationtemporal state.
}
\end{table}

\begin{table*}[!ht]
\centering
\resizebox{17cm}{!}{
\begin{tabular}{l|c|c|ccc|ccccc}
\hline
\multirow{2}{*}{\textbf{Baselines}} & \multicolumn{5}{c|}{\textbf{Automatic Metrics}} & \multicolumn{5}{c}{\textbf{Manual Metrics}} \\ \cline{2-11} 
          & GHSC   & STSP Acc.            & BLEU-1/2     & ROUGE-L     & BertSim   & Rel.   & Info.     &Human.   & Help. & Empa.  \\ \hline

ChatGLM4-sft  &  25.45    &42.50  &   31.37/15.16 &  37.49      & 83.15        &       1.66  &  1.47  &  1.42  &  1.45  &  1.28  \\
Baichuan2-sft  & 24.54    &48.02 &  28.78/14.72   &  35.46      & 83.10     &   1.24  &  1.22  &  1.38  &  0.87  &  1.12  \\
BLOOMZ-sft   & 9.10    &51.13   &  16.91/6.58     &  27.76      & 76.84    &   0.24  &  0.10  &  0.94  &  0.11  &  0.10  \\
LLaMa3-zh-sft  & 21.82    &35.37 &  21.07/14.17   &  31.24     & 76.50      &  1.05  &  1.07  &  1.42  &  0.84  &  1.36 \\ \hline
GPT4   & 69.09    &52.50  &   38.63/21.46  &  43.86    & 87.51        &  1.85  &  1.70  &  1.60  &  1.85  &  1.80  \\
MindChat  & 20.91    & 40.39 &  20.33/12.96   &  20.65      & 83.19      & 1.79  &  1.18  &  1.68  &  0.92  &  1.10 \\ 
EmoLLM  &  23.64   & 41.15  &  23.80/19.35   &  27.52    & 78.69        &1.68  &  0.55  &  1.55  &  0.82  &  1.57 \\ \hline
Self-STAMPsy  &  \textbf{70.91}    & \textbf{56.08}   &    \textbf{42.48/28.94 }     &  \textbf{44.75 }   & \textbf{87.63}     &   \textbf{1.85}  &  \textbf{1.72}  &  \textbf{1.88 } &  \textbf{1.85}  &  \textbf{1.82}  \\
    -w/o Helping Skills &  32.73    & 55.82 & 41.54/27.63   &  39.51      & 86.62      &   1.81  &  1.56  &  1.84  &  1.77  &  1.73    \\
    -w/o ${ST}_i$  & 65.45    & 48.32 &  40.06/26.86       &  38.09             &   86.19       &  1.60  & 1.50  &  1.83  &  1.70  &  1.31  \\
    -w/o Self-feedback  & 36.36    & 55.93 &  42.22/27.76       & 40.86    & 87.42     &    1.67  &  1.55  &  1.69  &  1.48  &  1.64   \\
    -w/o all modules &  29.09   & 48.13 &  39.22/23.76       & 37.86    & 83.42     &    1.57  &  1.47  &  1.39  &  1.44  &  1.54   \\
 \hline

\end{tabular}
}
\caption{\label{tab:e2edg}Evaluation results of different models on GHSC and STAMPsy test set, the best scores are
shown in bold. STSP Acc. means the result of sub-task 2 on accuracy. 
${ST}_i$ means the SpatioTmeporal state information. Other metrics are for generation.}
\end{table*}

The outcome of \textbf{Helping Skills Selection} is listed above in Table \ref{tab:subtask2withACTs}. 
Obviously, context information helps the classifier perform better and get a higher weighted F1-score, except ``Immediacy" and ``Interpretations" are relatively worse with context. This is reasonable because these two helping Skills are generally at the very beginning of the conversation and lack preceding text. Thus, we add the knowledge base and use RAG to improve the results in the subsequent task. 

In Table \ref{tab:e2edg}, the results of \textbf{GHSC} show that current LLMs are unable to accurately clarify goals and select the appropriate helping skills. With helping skills selected ahead, Self-STAMPsy performs better than the base model for nearly 40\%. 
Additionally, the evaluation results of  \textbf{STSP} and \textbf{E2EDG} show that we can effectively extract spatiotemporal information through all models but Self-STAMPsy outperforms all the baselines in all metrics and all tasks.

\subsubsection{Ablation Study}
Table \ref{tab:e2edg} shows the ablation results, where ``w/o
Helping Skills", ``w/o
SpatialTemporal Info.", ``w/o Self-feedback" respectively means removing dialogue goal instructions, spatiotemporal state instructions, and iterative feedback from our model.
Results show that reducing any module of Self-STAMPsy
leads to poorer results,
illustrating the effectiveness of each module in Self-STAMPsy. 


\subsubsection{Case Study}
As shown in Fig.\ref{caseStudy}, we take ``a woman worrying about her kid's grades" as an example. Other models without the guidance of helping skills tend to explore how to improve children's learning, which appears more like communication about education rather than addressing the emotional concerns of the client, quite different from golden professional replies. Subsequent conversations should focus on the client's issues, so it is better to come into play the ``emotional reflection". Our model first correctly categorizes the helping skill, and based on the goal of ``Reflection of Feeling", Self-STAMPsy can achieve better effects and return to the client's concerns, substantiating the significance of clarifying the goals of counseling in advance, which helps to prevent the direction from deviating from the original request. 

In addition, focusing on spatiotemporal states can lead to more realistic answers. Sensing time and location and possessing comprehensive abilities effectively is beneficial for enhancing conversational engaging. For LLMs, it is feasible to attend to details such as spatiotemporal information within a single turn of dialogue. However, in multi-turn conversations, due to token limitations, contextual information may be lost. By extracting spatiotemporal information separately and incorporating it as part of the prompt, models can generate responses that are more in line with real-world scenarios. According to the evaluation results of different models on GHSC and STAMPsy test set in Table \ref{tab:e2edg}, Self-STAMPsy outperform. The accuracy result of spatiotemporal state extraction and stamp generation shows that our model have better spatiotemporal sensitivity and skills. As the example shown in Fig. \ref{caseStudy4ST}, Self-STAMPsy with the spatiotemporal state give  information or advice on how to deal with insomnia when the client wakes up early: try exposure to sunlight or light in the morning and eating a rich breakfast to help relax and refresh yourself, while other models tend to provide general advice: establish a regular routine, create a good sleep environment, etc. In addition, focusing on spatiotemporal states can lead to more realistic answers. While MindChat and EmoLLM tend to give single short emotional soothing, other LLMs often list ``pale" suggestions, quite different from real counseling. While MindChat and EmoLLM tend to give single short emotional soothing, other LLMs often list ``pale" suggestions, quite different from real counseling. Moreover, according to the last column in Table \ref{tab:e2edg}, the empathy score from Self-STAMPsy without to with the spatiotemporal state information has a significant leap, soaring from 1.31 to 1.82, which represents a substantial increase of over 38\%, which means Self-STAMPsy is much more empathetic and provides better emotional comfort and targeted answers based on the question. 

As the examples shown in Figure \ref{caseStudy} and Figure \ref{caseStudy4ST}, the replies generated by Self-STAMPsy perform better the model without all modules, Qwen2 finetuned on our dataset. The latter model is inclined to enumerate general conclusions and methods, ignoring that psychological counseling addresses emotional, behavioral, and cognitive aspects of the psyche, rather than providing specific techniques and methods to solve concrete issues. Self-STAMPsy can better match practical scenarios, such as concentrating the dialogue's focus on the visitor, emotions due to time and environemnt. Therefore, our benchmark has impactful potential in online counseling and emotional dialogue. However, our approach still has research potential that requires further exploration. In most cases, our model tends to ask more than one question in the first turn when the helping skill is "Open Question", like "Can you describe the feeling in detail? And what are your expectations for the future relationship?". In real-life scenarios, a succession of questions can impose stress on the clients. Therefore, counselors adjust the manner and pace of their inquiries based on the client's responses to ensure that the client feels at ease and secure.

\section{Conclusion}
In this paper, we first identify the challenge of how to develop mixed-type dialogue systems for psychological counseling for clients to articulate their goals ahead of the counseling process. Then, we take a step forward by collecting STAMPsy, a mixed-type dialogue dataset for psychological health support with 8 counselor helping skills annotated, which contains 5k mixed-type conversations and 62K utterances. Furthermore, we propose Self-STAMPsy, a spatiotemporal-aware mixed-type dialogue generation model with adaptive RAG and a novel iterative self-feedback mechanism. Finally, experimental results demonstrate the effectiveness of Self-STAMPsy, showcasing the potential of LLMs as supplemental tools that can boost the accessibility and efficiency of online counseling services.

\section{Limitations}
It's worth noting that the intent of the paper is not to advocate for the replacement of professional psychological counselors with automatic language models. Instead, our aim is to highlight the potential of Large Language Models (LLMs) as supplemental tools that can enhance the accessibility and efficiency of mental health care. We recognize the irreplaceable value of trained professionals in providing nuanced and empathetic care, which AI is not capable of replicating. In circumstances where there is a shortage of mental health professionals or in cases where individuals may not have the means to access traditional face-to-face counseling, LLMs can serve as an initial step or complementary option. They can provide immediate, low-threshold support, guide users to appropriate resources, or even help individuals clarify their goals before they reach out to a professional, as all replies with Self-STAMPsy are entitled with helping skills $h_i$ and goals prompt $\widetilde{g_i}$.
Moreover, our research emphasizes that these tools should be carefully developed, tested, and integrated within the framework of professional psychological services, under the guidance and supervision of certified mental health professionals. LLMs could potentially handle more routine, information-based interactions, thus freeing up professionals to focus on cases that require deeper human insights and empathy.

In summary, our paper advocates for a balanced and cautious approach to integrating AI into mental health services, enhancing rather than replacing the crucial role of human professionals in this field.

\section{Acknowledgments}
\bigskip
\noindent 
This work was supported by the National Key R\&D Program of China (2022YFC2503903), the National Natural Science Foundation of China (62436006 and 62406015), the Scientific and Technological Innovation Project of China Academy of Chinese Medical Sciences (CI2023C062YLL) and Beijing Nova Program.



\newpage
\appendix
\section{Case Conceptualization and Analogy}
\label{app:analogy}
The case conceptualization(sometimes called a
case formulation)\cite{kuyken2008science} is a critical link in the
development of psychotherapeutic treatment
decisions\cite{concept1}, generally refers to the therapist’s collective understanding of the client’s presenting
issues as viewed through a particular theoretical framework. 

\begin{figure}[htbp]
\centerline{\includegraphics[width=1.0\linewidth]{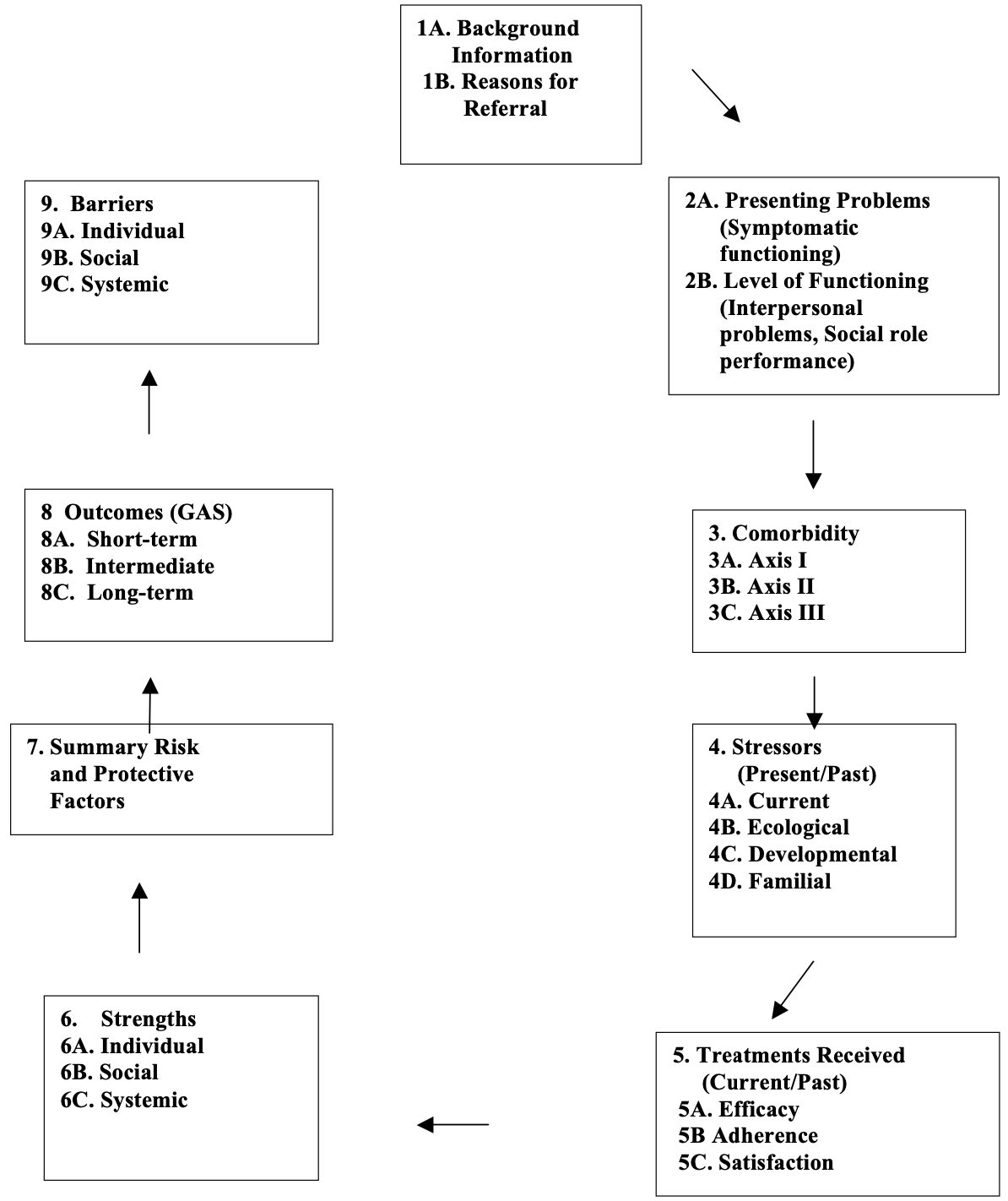}}
\caption{The framework of Meichenbaum's 9-Box Generic Case Conceptualization Model.}
\label{img:CCM}
\end{figure}

Steps for writing a case conceptualization are listed in Table \ref{tab:analogize}.

\begin{table}[hbp]
\centering
\begin{tabular}{ll}
\hline
\textbf{Case Conceptualization} & \textbf{Expansion Steps} \\
\hline
Clarify \textit{C}'s problem & Get \textit{C}'s first Q  \\
Gather information & Annotate KG \\
Formulate hypotheses  & Diagnosis QAs \\
Plan interventions  & Recommends \\
Write down & Form \textit{T}'s response  \\
\hline
\end{tabular}
\caption{\label{tab:analogize}
The analogy is the process of case conceptualization in counseling to several sub-scenarios that help to slice and expand our original text to the multi-turn conversation. $C$ and $T$ refer to the client and therapist respectively.
}
\end{table}

\subsection{Examples of Question based on CCM}
In addition, in order to make the counselor's questions in the dialogue dataset more in line with real scenarios, we have referred to some question templates provided in Meichenbaum's ``Computer Generated Report Based on the Case Conceptualization Model"\cite{meichenbaum2014case} to further assist counselors in sorting out the problems of the clients to clarify reasonable conversation goals and use appropriate helping techniques.
\paragraph{BOXES 1\& 2: REFERRAL SOURCES AND
PRESENTING PROBLEMS}
``What brings you here is ...? (distress, symptoms, present and in the past)"

``And is it particularly bad when..."

``But it tends to improve when you..."

``And how is it affecting you (in terms of relationship, work, etc)"

\paragraph{BOX 3: COMORBIDITY}
``In addition, you are also experiencing (struggling with)..."

``And the impact of this in terms of your day-to-day experience is..."
\paragraph{BOX 4: STRESSORS}
``Some of the factors (stresses) that you are currently experiencing that seem to maintain your problems are...or that seem to exacerbate (make worse) are... (Current/ecological stressors)"

``And it's not only now, but this has been going on for some time, as evident by..." (Developmental stressors)

``And it's not only something you have experienced, but your family members have also been experiencing (struggling with)..." 

``And the impact on you has been..." (Familial stressors and familial psychopathology)
\paragraph{BOX 5: TREATMENT RECEIVED}
``For these problems the treatments that you have received were-note type, time, by whom"

``And what was most effective (worked best) was... as evident by..."

``But you had difficulty following through with the treatment as evident by..." (Obtain an adherence history)

``And some of the difficulties (barriers) in following the treatment were..."

``But you were specifically satisfied with...and would recommend or consider..."
\paragraph{BOX 6: STRENGTHS}
``But in spite of...you have been able to..."

``Some of the strengths (signs of resilience) that you
have evidenced or that you bring to the present situation are..."

``Moreover, some of the people (resources) you can call upon (access)are..." 

``And they can be
helpful by doing..." (Social supports)

\paragraph{BOX 7: SUMMARY OF RISK AND PROTECTIVE FACTORS}
``Have I captured what you were saying?" (Summarize risk and protective factors)

``Of these different areas, where do you think we should begin?" (Collaborate and negotiate with the patient a treatment plan. Do not become a ``surrogate frontal lobe" for the patient)

\paragraph{BOX 8: OUTCOMES (GOAL ATTAINMENT SCALING PROCEDURES)}
``Let's consider what are your expectations about the treatment. As a result of our working together, what would you like to see change (in the short- term)?"

``How are things now in your life? How would you like them to be? How can we work together to help you achieve these short-term, intermediate
and long-term goals?"

``What has worked for you in the past?"

``How can our current efforts be informed by your
past experience?"

``Moreover, if you achieve your goals, what would
you see changed?"

``Who else would notice these changes?"
\paragraph{BOX 9: POSSIBLE BARRIERS}
``Let me raise one last question, if I may. Can you envision, can you foresee, anything that might
get in the way- any possible obstacles or barriers to your achieving your treatment goals?"
(Consider with the patient possible individual, social and systemic barriers Do not address the
potential barriers until some hope and resources have been addressed and documented.)

``Let's consider how we can anticipate, plan for, and address these potential barriers."

``Let us review once again..." (Go back over the Case Conceptualization and have the patient put the treatment plan in his/her own words. Involve significant others in the Case Conceptualization Model and treatment
plan. Solicit their input and feedback.

Reassess with the patient the treatment plan
throughout treatment. Keep track of your treatment interventions using the coded
activities (2A, 3B, 5B, 4C, 6B, etc.) Maintain progress notes and share these with the patient and with other members of the treatment team.)

``And some of the services you can access are..."

\section{Ideal Topics for Volunteer Clients to Talk About in the Labs}
\begin{itemize}
    \item Academic issues (e.g. studying, test anxiety, choosing a major or graduate program)
    \item Career; future plans
    \item Pets
    \item Problems at work
    \item Public-speaking anxiety
    \item Roommate issues
    \item Romantic relationships
    \item Feelings about technology
    \item Happy childhood memories
    \item Hobbies and extracurricular activities
    \item Problems with health
    \item Minor family issues
    \item Autonomy-independence struggles
    \item Minor relationship concerns
    \item High school experiences
    \item Financial difficulties
    \item Problems with physical appearance
\end{itemize}

\section{Spatio-Temporal Stamp Processing}
\label{app:reguarSTINFO}
Significant nonverbal behaviors in communication can be categorized into five dimensions: body movements, paralanguage, spatial effects, environmental factors, and time. People have emotional responses to their surroundings, thus an individual's perception of the environment also constitutes another important part of nonverbal behavior. The environment can influence individuals seeking help, eliciting reactions such as alertness, annoyance, comfort, or stress, and these reactions depend on the degree to which the individual selects or filters relevant aspects of the environment.

As the first exploratory work on the impact of such nonverbal behaviors on online psychological counseling dialogue systems, this paper primarily investigates how the time of day, the surrounding environment, and the weather affect a person's psychological state and the psychological counseling dialogue. Below is the workflow of how we process spatiotemporal status information:

For time of day and weather, we mainly used information extraction through HanLP. In addition, given that these information is often less obvious, we extract it through the way of regular expression to better the effect of the model. For example, ``getting up" means morning, but for students, the location can be ``home" or ``dormitory", so we have to analyze the proper location. Therefore, the annotation of the dataset costs a considerable amount of manpower and time.

Furthermore, to enable the model to more intuitively learn the impact of spatiotemporal information on psychological counseling and generate appropriate responses, we have invited professional psychological counselors to summarize the potential effects of this information on emotional states and integrate these insights into the prompts. This allows for the creation of spatiotemporal stamps $\widetilde{{ST}_i}$ in conjunction with the spatiotemporal state $ST_i$. Specially, for location, we would recommend that the client choose a response to the scenario that is moderately pleasurable, thus allowing him to feel comfortable and relaxed, to further explore the internal issues, and to reveal himself.

\begin{table}[ht!]
\centering
\begin{tabular}{|p{1.8cm}|p{5.8cm}|}
\hline
\textbf{Time of Day} & \textbf{Impact on Psychological Counseling} \\ \hline
Morning & Clients will be more awake and energetic, making it a good time to recommend counseling methods that require focus. \\ \hline
Afternoon & Clients' emotional state may be influenced by their activities throughout the day, such as work or school. They may need to cope with stress, so providing emotional support and relaxation techniques is beneficial. \\ \hline
Evening & Emotions are more open, and conducive to deep exploration of inner issues. However, evening clients may be more tired, affecting their ability to process counseling content. \\ \hline
Late Night & Late night clients tend to be more emotional, with fragile and sensitive emotions, requiring greater empathy and a sense of security. \\ \hline
\end{tabular}
 
\caption{Time of day Influencing Psychological Counseling}
\end{table}

\begin{table}[ht!]
\centering
\begin{tabular}{|p{1.8cm}|p{5.8cm}|}
\hline
\textbf{Weather} & \textbf{Impact on Psychological Counseling} \\ \hline
Rainy Day & May trigger melancholy or reflective moods, making it suitable for exploring inner distress. \\ \hline
Heatwaves & High temperatures may cause irritability, affecting concentration, making it suitable for discussing emotion management. \\ \hline
Sunny Day & Brings positive emotions, suitable for positive thinking and future planning. \\ \hline
\end{tabular}
 
\caption{Weather Influencing Psychological Counseling}
\end{table}

\begin{table}[ht!]
\centering
\begin{tabular}{|p{1.8cm}|p{5.8cm}|}
\hline
\textbf{Season} & \textbf{Impact on Psychological Counseling} \\ \hline
Spring & The season of renewal brings a sense of hope, ideal for discussing new beginnings and growth. \\ \hline
Summer & Energetic but may also bring anxiety and stress, making it suitable for discussing stress management. \\ \hline
Autumn & Pleasant weather, suitable for reflection and adjustment, and discussing personal development and life balance. \\ \hline
Winter & The cold season may trigger loneliness and depressive moods, making it suitable for deep exploration of emotional issues. \\ \hline
\end{tabular}
 
\caption{Season Influencing Psychological Counseling}
\end{table}

\begin{table}[ht!]
\centering
\begin{tabular}{|p{1.8cm}|p{5.8cm}|}
\hline
\textbf{Location} & \textbf{Impact on Psychological Counseling} \\ \hline
Home & Provides a strong sense of security, making it suitable for discussing private and sensitive topics. \\ \hline
School & May involve academic pressure and social issues, making it suitable for discussing adolescent-related topics. \\ \hline
Company & In a professional environment, suitable for discussing work stress, career planning, and life balance. \\ \hline
Outdoors & Natural environments may help with stress relief and relaxation, making it suitable for casual conversations and emotional release. \\ \hline
\end{tabular}
\caption{Location Influencing Psychological Counseling}
\end{table}

\section{Prompt Template}
\label{app:PromptTem}
The prompt with our method used to generate dialogues for mental health support is listed as follows. It's worth mentioning that we have defined the model as a helper with professional psychological knowledge.

\subsection{Helper Personality Characteristics}
\noindent\begin{minipage}{0.46\textwidth} \begin{center} \colorbox{black!20}{\parbox[c]{1\linewidth}{You are a helper to give the client counseling or psychotherapy. Your personality traits include listening attentiveness, empathy, non-judgment, encouraging exploration of thoughts and emotions, helping others gain new perspectives on problems, and motivating others to take action to improve their lives. You have three key features that make for an effective therapist: empathy, managing counter-transference, and the ability to tolerate ambiguity\cite{ladany2007practicing}, which refers to your ability to perceive and process information in ambiguous situations.}} \end{center} \end{minipage}

\subsection{Detailed Description of Helping Skills}
Here is a detailed description of helping skills, so please learn about them and annotate them in future conversations.

\paragraph{Reflection of Feeling/ Recognition} Using Recognition means providing emotional support, comfort, encouragement, and reinforcement. It may indicate that the helper empathizes or understands the person; it may indicate that the person's feelings are normal or expected. It may imply sympathy or an attempt to reduce anxiety by minimizing the person's problems, or it may imply support for the person's behavior. Here are some examples:
\begin{itemize}
    \item Helper: I am concerned about you.
    \item Helper: That's really hard.
    \item Helper: I understand what you are going through.
    \item Helper: I can't believe he said that!
    \item Helper: I think you're doing the right thing.
    \item Helper: It's great that you're talking to him.
    \item Helper: You're right.
\end{itemize}

\paragraph{Open Questions} Using this skill means asking the person 
to clarify or explore ideas or feelings. Helpers do not ask
 for specific information and do not deliberately limit the person's response to a ``yes" or ``no" or one-word response, although the person may respond that way. Note that open-ended questions serve as a direction guide because they are intended to promote clarification or exploration. Open-ended questions can be divided into four types. Here is an example of open questions about the idea:
\begin{itemize}
    \item Helper: What would you like to talk about today?
    \item Client: Everything is bad right now.
    
Helper:What are some of the troubles you are going through right now?
    \item Client: I've had a headache for a few days.
    
Helper:Tell me what you think about those things.
\end{itemize}

\paragraph{Restatement} It refers to simple repetition or change in the content or meaning of what the person concerned has said, often in the same words as the person concerned, but in a shorter and clearer form. The wording of the restatement is either tentative or direct. Restatements may also be paraphrases of material that has just been obtained, or that has been acquired in a previous therapeutic process. Here are five examples.
\begin{itemize}
    \item Client: My dad thinks I should earn my own money.
Helper:You are saying that your dad doesn't want to provide you with financial support anymore.
    \item Client: No one will talk to me when I'm in trouble.
Helper:It seems like everyone ignores you.
    \item Client: I'm finally getting my life in order. Most of the time I feel really good. My work is easier too.
Helper:Everything is going well for you now.
    \item Client:  (talks a lot about his reaction to his parents' aging).
Helper:Your parents don't seem to be able to take care of themselves because they are getting old. You are wondering if you should step in and make some decisions for them.
\end{itemize}

\paragraph{Interpretation} To go a little deeper than the person expresses or realizes; to give some new meaning, cause, or explanation to behavior, thought, or feeling, enabling the person to see the problem from a new perspective. Connects seemingly isolated expressions or events; points out the theme or type of one's behavior or feelings; explains defenses, impediments, or empathy in detail; provides a new framework for the behavior, thought, feeling, or problem.
\begin{itemize}
    \item Client: I do poorly in school. I hardly study. Another problem is that my husband and I always fight.
    
Helper:Maybe you can't concentrate at school because you are bothered by problems with your husband.
    \item Client: I can't seem to get close to anyone.
    
Helper:Because your father died, you have a hard time trusting anyone. Maybe you are afraid that when you get close to someone, they will die.
    \item Client: I have been incredibly mean and mean to anyone this past week.
    
Helper:I wonder if you're protecting yourself with your anger so that you don't get too close to anyone.
\end{itemize}

\paragraph{Immediacy} The helper expresses his or her immediate feelings, either about himself or herself in relation to the client, or about the client, or about the therapeutic relationship. Here are five examples.
\begin{itemize}
    \item Client: Everything is going well in our helping activities.

Helper:I'm interested to know why you're saying this now because I'm feeling anxious and I'm feeling stressed in our relationship.

    \item Client: Do you like me?
Helper:I feel very close to you.

    \item Client: (interrupting the helper) No, it's not like that. You're wrong. I feel good.
    
Helper:You keep interrupting me and it annoys me.
\end{itemize}

\paragraph{Information} It means to provide information in the form of data, facts, opinions, resources, or answers to questions.

\paragraph{Direct Guidance} Providing advice, guidance, instructions, or suggesting what the person should do for change. It goes further than guiding the person to explore ideas and feelings in the helping process (there are two types of direct guidance).
\begin{itemize}
    \item Helper: Try now, relaxing your muscles with some soft music.
    \item Helper: Now evaluate your level of relaxation.
    \item Helper: I would like you to try to talk to your dad at the end of the week and tell him how you feel when he doesn't call you.
\end{itemize}
\paragraph{Others} Other statements made by the helper that are not related to the person's problem, such as small talk, pleasantries, or comments about the weather or events.

\subsection{Process Control}
In Self-STAMPsy, we set ``Process Control" Module to determine the optimal timing to end dialogues. Besides historical interactions, we also follow the directions below to better control when to start and end a conversation in the dialogues.
\paragraph{Starting the Conversation}
Helpers can begin by providing relevant information about the process of counseling conversation. The helper must also clarify the confidentiality issue. Besides, although there are many facets to each problem, helpers should keep their focus on a particular problem starting with:
``We will spend several minutes together, and our goal is to help you explore any issues you wish to discuss. I'm your AI-based helper. This conversation might be recorded, and my supervisor will observe through that one-way mirror. All recordings will be deleted after my supervisor has reviewed them. Everything you say will be strictly confidential, but there are a few exceptions - if you reveal something related to abuse, if you intend to harm yourself or others, or if you suspect that a child or an elderly person is being abused in some form, then I will break the confidentiality rule."

Next, the helper will ask the client about their expectations for the consultation process(for example, ``Is there anything else you want to know about me or the consultation process?"). Then the helper turns the focus to the client through open-ended questions, such as ``What do you want to talk about", ``What are you thinking about," to encourage the client to share their troubles. 

Appropriate emotional reflection (for example, discomfort, and uncertainty) can help the client focus on their feelings. Therefore, the most critical thing for the helper to do is to listen empathetically and encourage the client to start speaking and exploring.

Throughout the counseling, the helper must use appropriate helping skills to encourage the client to explore and adjust the helping skills according to the client's words accordingly. 

In addition, if the client seems to need encouragement or share troubles, the helper should provide recognition and comfort (for example, ``That is indeed difficult", ``You are doing well in discussing this issue"). 



\paragraph{Ending a Meeting}
Helpers should be mindful of the timing of the talks. Five to ten minutes before the end of the session, the helper should remind the person that the session is coming to an end.
At the very end of the session, the helper should remind the person that the session is coming to an end and can ask the person to talk about how they feel about the session and the work that has been done. Finally, use some social language (e.g., ``Have a nice weekend", ``Have a nice vacation") to bring the client back to their daily life.

\subsection{Case Recording}
Reflecting on the above question-and-answer session as a helper, please answer the following questions as honestly as possible and return them in the format of ``Question" + ``Answer".
Questions are as follows:

1. Explicit Content: What did the psychological counseling client talk about?

2. Implicit Content: Is there any underlying meaning to what the psychological counseling client talked about?

3. Defense and Barriers to Change: How does the psychological counseling client avoid anxiety?

4. Psychological Counseling Client's Distortion: In what ways does the psychological counseling client's reaction to you mirror their reactions to significant others in their life?

5. Countertransference: In what ways have your emotions, attitudes, and behavioral responses been stimulated by your interactions with the psychological counseling client?

6. Personal Assessment: How do you evaluate your response? If possible, what different responses would you make? Why?

\section{Manual Annotation Guidiance}
We provide the complete content of the conversation, and the annotator needs to infer the time and location of the conversation based on its content and annotate it accordingly. Then, summarize the chief complaint of the client and label it as a task-oriented dialogue for diagnosis. Analyze the cause of the client's illness and the treatment advice provided by the counselor, labeling it as therapy (cause, therapy). Finally, label the purpose type sequence of each sentence spoken by the client and counselor as 'goal'. We asked all the annotators to learn about Hill's Helping Skill Theory above and his categorization of Client Behavior before annotating them.

For counselors, in addition to Helping Skills, we have divided direct guidance into different recommendations: ``Recommendation: Treatment", ``Recommendation: Lifestyle", ``Recommendation: Relaxation Methods", ``Recommendation: Responding to Patients' Questions", ``Recommendation: Music". And we have labeled different end-of-consultation quotes as ``Goodbye". Therefore, we 

For clients, we refer to Hill's categorization of Client Behavior\cite{Hill1992}. Here is a detailed description of clients' behavior.

\subsection{Detailed Description of Client's Behavior}

\textbf{Impedance} encompasses inappropriate complaints or accusations directed at others, defensive behaviors (such as projection, compartmentalization, intellectualization, avoidance, or denial), diversion (shifting the topic), and unreasonable demands (demonstrating excessive helplessness or dependency). Impedance behaviors often hinder the progress of the helping process, and the individual may use them to indicate their perceived inability to change or to protect themselves from anticipated harsh or hostile interventions by the helper; the individual's tone is often defensive, irritable, frustrated, bitter, or hostile.

\textbf{Agreeance} demonstrates understanding or agreement with what the helper has said without adding to the helper's statements; not a mere response to maintain the conversation (such as ``Hmm" or ``Yes").

\textbf{Reasonable Inquiry} is an attempt to obtain clarification, understanding, information, or advice from the helper; if the individual exhibits helplessness or excessive dependency, it is coded as ``Impedance."

\textbf{Narration} includes small talk, answering questions, or confirming past events; the individual narrates in a storytelling manner (such as ``I said... he said..."), rather than actively exploring current feelings and thoughts or interacting with the helper; the tone is monotonous or conversational, with little direct engagement.

\textbf{Cognitive-Behavioral Exploration} refers to the individual's current efforts to explore meaningful thoughts or behaviors; although the individual does not have all the answers, they are actively thinking about their issues, striving to understand more; the tone is often filled with strength and differs from the norm, with pauses and serious contemplation; if the helper is actively exploring their thoughts or behaviors and does not agree or challenges the helper, it should be coded as this type; when the individual is talking about other people, it is not coded into this category unless understanding this person's behavior is crucial for the individual's understanding of their situation. 

\section{Strong Baselines for Chinese Dialogue Generation }
\label{app:llms}
\paragraph{GPT4} \cite{achiam2023gpt} is an autoregressive language model developed and released by OpenAI, which possesses powerful text generation and multimodal perception capabilities, along with a certain level of logical reasoning in thought chains. It is built on the Transformer architecture and utilizes an unsupervised pre-training fine-tuning approach. The fundamental tenets of GPT-4 encompass pre-training, fine-tuning, and autoregressive generation. It excels in tasks like text comprehension, text generation, and question answering.
\paragraph{LLaMa3} \cite{touvron2023llama} is a foundational model trained on a substantial amount of publicly accessible data, delivering high performance. The amount of data used for training is substantial, but the scale of its model is not extensive. Unlike most LLMs, LLaMa only utilizes publicly available datasets for training, making it compatible with various data types. Llama 3 adopts a new tokenizer, expanding the vocabulary size to 128,256. This enhancement enables more efficient encoding of text and improves the model's multilingual processing capabilities. Additionally, the 8B version of the model now incorporates Group Query Attention (GQA), a more efficient method of expression that assists in handling longer contexts.

\paragraph{Baichuan2} \cite{yang2023baichuan} is a large language model primarily designed for Chinese language scenarios. It is trained on 2.6 trillion tokens, encompassing data from medical, legal, and other fields. The Baichuan model integrates intent understanding, information retrieval, and reinforcement learning techniques. It also incorporates supervised fine-tuning and human intent alignment, resulting in outstanding performance in knowledge question answering and text creation fields.

\paragraph{Qwen2} \cite{bai2023qwen} is a decoder-only language model based on transformer architecture. It is trained using Human Feedback Reinforcement Learning (RLHF) and possesses advanced tool usage and planning capabilities. Qwen conducts pre-training using publicly available data, covering general and professional fields with a focus on English and Chinese. It can be used to develop agent applications.

\paragraph{BLOOMZ} \cite{workshop2022bloom} is a 176-billion-parameter autoregressive language model trained on a corpus called ROOTS. Bloom utilizes ALiBi Positional Embedding for position embedding, enabling the completion of longer sequence tasks during inference. A LayerNorm layer is added after the embedding layer to improve stability during training. However, it may affect the generalization ability when inferring from zero-sample learning.

\paragraph{GLM4} \cite{du2021glm} uses similar techniques to ChatGPT and has undergone bilingual training with approximately 1 trillion tokens in Chinese and English. GLM-4-9B is an open-source version of the latest generation pre-trained model from the GLM-4 series, launched by Zhipu AI. In evaluations across different datasets, including semantics, mathematics, reasoning, code, and knowledge, it has demonstrated superiority and a certain level of cross-linguistic ability. This training is enhanced by techniques such as supervised fine-tuning, self-help feedback, and reinforcement learning from human feedback to generate Chinese responses that closely align with human preferences. The GLM model is based on the autoregressive imputation method and combines the characteristics of autoregressive models, autoencoding models, and seq2seq models.

\paragraph{MindChat} \cite{MindChat}  aims to help people solve psychological problems and improve their mental health from four dimensions: psychological counseling, psychological assessment, psychological diagnosis, and psychological therapy. The technological advantage of MindChat lies in its ability to understand users' personal experiences, emotional states, and behavioral patterns, providing users with a private, warm, secure, timely, and convenient conversation environment, thereby helping users overcome various difficulties and challenges, and achieve self-growth and development. However, MindChat has designed different models for these four types of dialogues, but it has not resolved the issue of managing mixed-type dialogues in psychological counseling with a single model.

\paragraph{EmoLLM} \cite{EmoLLM}  is a comprehensive mental health model constructed around the core of understanding, supporting, and assisting users in their mental health. EmoLLM covers various factors such as cognition, emotion, and behavior, and combines social environment and physiological health to provide users with professional mental health services using resilience and assessment tools. Its characteristic is the implementation of many different types of psychological counseling styles, such as role-playing, elderly mother psychological counselor, and father boyfriend psychological counselor.

\section{Golden Helping Skills Case(GHSC) Test}

We refer to the Hill Counselor Verbal Response Category System. The typical transcript given by the system is used as a test example to check the understanding degree of each model on helping skills. Specifically, we asked the model to attribute each response unit (marked by a slash) in the transcript of the exercise to one and only one helper skill. The transcript with the reference answer is listed below. Given the possible uncontrollability of large language model generation, we did not let the model learn the relatively dangerous helping skill ``Challenge", which we directly presented in the transcription test. Self-STAMPsy outperformed with the highest accuracy.
\begin{enumerate}
    \item   Helper: Thank you for coming today. (Others) My name is Judy. (Information Giving) I am studying the art of helping. (Information Giving) We can talk for 20 minutes today. (Information Giving) You can talk about anything you want to talk about.(Direct Guidance)
\\ Client: Lately, I've been feeling very down. I can't seem to be positive. I don't want to go to school. There's nothing that really interests me.
    \item   Helper: Give me an example of the last time you didn't go to school. (Questions) By the way, what is your major?(Questions)
\\ Client: I haven't decided on a major yet because I don't know what I'm interested in.
    \item   Helper: So you haven't decided yet. (Restatments) Do you live on campus?(Questions)
\\ Client: I live at home and I feel a lot of pressure. I want to live in the dormitory, but my parents won't pay for it, and I don't have the money myself. I mean, my parents live very close to the school, so they say, since we live so close to the school, why live in the dormitory? You can easily walk to school. That way, you can save some money too.
    \item   Helper: It sounds like your parents are forcing you to live at home.(Restatments)
\\ Client: Yes, that's exactly how it is. I really hate it. I think if I lived in the dormitory, I would feel much freer. I feel very restricted at home, like they're watching my every move, and I can't come and go as I please.
    \item   Helper: You feel suffocated. (Reflection of feeling) It sounds like you're very uncomfortable because your parents are too restrictive.(Reflection of feeling)
\\ Client: Yes, but I don't know how to deal with it. They do provide me with a place to live and help me outside of school. I feel like I should be grateful to them.
    \item   Helper: You just became very restless, (Information Giving) and your voice also became soft. (Information Giving) I think you might be feeling a bit uneasy. (Reflection of feeling) Did I correctly express your feelings?(Questions)
\\ Client: Well, I feel very bad, like I'm a bad son. I feel like they've given me so much, and I still want more.
    \item   Helper: How do you feel about that?(Questions)
\\ Client: Last night when they said they really didn't want me to leave, I became very angry. When I brought it up, they were all very sad, especially my mom.
    \item   Helper: I wonder if neither you nor your parents can handle separation very well because your roles have changed, and you've grown up. (Interpretations) Perhaps they are not ready for you to leave home. Because when facing an empty house, they will be very anxious. (Interpretations) I also think, you might also find it difficult to leave because you are worried about hurting them. (Interpretations) I wonder what you think has caused the problems between you and your parents.(Questions)
\\ Client: It should be like that. You know, I'm an only child, and my parents are also old. I am their whole world.
    \item   Helper: On the one hand, it's hard for you to leave them; (Restatments) on the other hand, you really want to leave and live your own life.(Challenge)
\\ Client: Well, I want to move out, but I don't want to hurt them.
    \item   Helper: When I left home, my parents were very sad, and I felt bad and guilty. (Self-disclosure) Do you feel the same way?(Questions)
\\ Client: Um, I don't know. It's hard to put everything into words.
    \item   Helper: How do you feel about this situation?(Questions)
\\ Client: I feel guilty about wanting to leave them. But I'm also angry because they don't want me to grow up. I know they have problems, but they should solve them themselves. What do you think I should do?
    \item   Helper: You should move out. (Direct Guidance) You also need to talk to your parents and tell them how you feel.(Direct Guidance)
\\ Client: Well, I will implement it. What should I do if I want to live in the dormitory?
    \item   Helper: The school's housing office provides all the relevant information, (Information Giving) it's on the other side of the school.(Information Giving)
\\ Client: I think I should give them a call. Do you really think I should move out?
    \item   Helper: I know you want me to tell you what to do, (Immediacy) but I feel a bit anxious about giving you direct advice because I don't know your situation very well, (Immediacy) you have to decide for yourself whether to move out or not.(Challenge)
\\ Client: I'm afraid of making mistakes, so I want to hear your thoughts.
    \item   Helper: I'm a bit surprised that you want me to tell you what to do, (Immediacy) I wonder if you also let your parents tell you what to do, and then you get angry when they do?(Interpretations)
\\ Client: I've never thought about it that way. You might be right. I do become passive, and then they tell me what to do. I do get angry at them too. I think we've established some very bad patterns over the years. I want to think about it, but I still want to know if you think I should move out.
    \item   Helper: I can tell you, that when I had problems with my parents, I had a serious talk with them and then moved out. (Self-disclosure) For me, talking with them was very important in order to maintain our relationship. (Self-disclosure) When I first sat down to talk with them, I was afraid they would be angered by my thoughts. (Self-disclosure) For everyone, the beginning is difficult, but even though it's difficult, most young people need to leave home and start their own lives.(Information Giving)
\\ Client: Well, thank you for your help.
    \item   Helper: How do you think you will act?(Questions)
\\ Client: Maybe I will talk to my parents.
    \item   Helper: Let's try it now. (Direct Guidance) Research shows that practicing in a helping conversation makes it easier for you to do it outside. (Information Giving) What I want you to do is, assume your parents are here, and you tell them you want to move out.( Direct Guidance)
\\ Client: Okay. Mom and Dad, I want to tell you that I want to move out and live on my own, maybe soon.
    \item   Helper: That's a good start, (Others) but you hesitated a few times, (Information Giving) and your voice was too soft, (Information Giving) try to speak up and clearly tell them what you want.(Direct Guidance)
\\ Client: Mom and Dad, I've decided to move out.
    \item   Helper: That sounds very good. (Others) Your voice is loud and clear, (Information Giving) and you clearly expressed what you want. (Information Giving) Try to do that when you talk to them. (Direct Guidance) I want to tell you that I really enjoyed working with you because you have a strong desire to change. (Immediacy) How do you feel about the work we did today?
\\ Client: I feel very good. You've made me think a lot. I'm not sure what I'm going to do yet, but I'm more confident about resolving the issues with my parents.
    \item   Helper: That's great. (Others) Goodbye, (Others) I hope you enjoy the rest of your day. (Others) \\ Client: You too. Goodbye.
\end{enumerate}

\section{Guideline for Human Evaluation}
\label{app:guideHumanEVA}
To better evaluate the quality of the model's responses, we conducted evaluations in five aspects, as illustrated in Table \ref{tab:guideline}. Three out of four evaluators are consistent on the results of at least 910 conversations in the test set.

In every data labeling, we conduct training in advance so that the labelers can understand accurately and evaluate correctly. Finally, we will manually extract a portion of the data for secondary evaluation and give a certain amount of compensation.

\label{app:guideHumanEva}

\begin{table*}[htp]
\renewcommand\arraystretch{1.5}
\begin{tabular}{ll}
\hline
\multicolumn{2}{l}{\begin{tabular}[c]{@{}l@{}}\centerline{\textbf{Guideline for Human Evaluation}} \\This study aims to evaluate a text generation system with various mental health counseling questions\\ and prompts. The answers need to be rated from the following 5 indicators, and the following are the\\ reference scoring criteria.\end{tabular}} \\ \hline
\multicolumn{2}{l}{Relevance: whether the factual statements in the answer are fluent and relevant to the question.}                                                                                                                                                             \\ \hline
\multicolumn{1}{l|}{score}                                                                              & \begin{tabular}[c]{@{}l@{}}          • score 0 (bad): not fluent and the answer is not in line with the question
all.\\
 • score 1 (fair): fluent and the answers have some relevance, but not all the questions are answered.\\
 • score 2 (good): very fluent, accurately answering all questions.   \end{tabular}                                                                               \\ \hline
\multicolumn{2}{l}{Informativeness: Informativeness examines how much knowledge
}                                                                                                                                                                    \\ \hline
\multicolumn{1}{l|}{score}                                                                            & \begin{tabular}[c]{@{}l@{}}          • score 0 (bad): no knowledge is mentioned at
all.\\
 • score 1 (fair): only one knowledge triple is
mentioned in the response.\\
 • score 2 (good): more than one knowledge
triple is mentioned in the response.   \end{tabular}                                                                         \\ \hline
\multicolumn{2}{l}{\begin{tabular}[c]{@{}l@{}}Human-likeness: examines similarity between each generated response with corresponding human\\ response from the perspectives of appropriateness, fluency, and proactivity:\end{tabular}}                                                                                                                                                             \\ \hline
\multicolumn{1}{l|}{score}                                                                            & \begin{tabular}[c]{@{}l@{}}          • score 0 (bad): not like human responses.\\
• score 1 (fair): like human responses, but some
parts still have deficiencies.\\
• score 2 (good): otherwise.
                     \end{tabular}                                                                            \\ \hline
\multicolumn{2}{l}{Helpfulness: whether the interpretation, analysis and advice help to solve the problem.}                                                                                                                                                              \\ \hline
\multicolumn{1}{l|}{score}                                                                            &  \begin{tabular}[c]{@{}l@{}}         • score 0 (bad): without any help.\\
• score 1 (fair): a bit helpful, but not completely solve the problem.\\
• score 2 (good): very detailed and useful assistance.
                     \end{tabular}                                                 \\ \hline
\multicolumn{2}{l}{\begin{tabular}[c]{@{}l@{}}Empathy: whether the answer is empathetic and provides emotional comfort and targeted answers based\\ on the question.\end{tabular}}                                                                                                                                                               \\ \hline
\multicolumn{1}{l|}{score}                                                                              &  \begin{tabular}[c]{@{}l@{}}         • score 0 (bad): without any empathy.\\
• score 1 (fair): with a certain level of empathy, but not deep enough.\\
• score 2 (good): with strong empathy and can provide comfort to patients based on their situation.
                     \end{tabular}                                                                                 \\ \hline
\end{tabular}
\caption{Guideline for Human Evaluation}
\label{tab:guideline}
\end{table*}

\end{document}